\documentclass[journal]{IEEEtran}
\usepackage{graphicx,subfigure}
\usepackage{epstopdf}
\usepackage{amsmath,amssymb}
\usepackage{bm}
\usepackage{color}
\usepackage{multirow}

\begin{document}

\title{ReMarNet: Conjoint Relation and Margin Learning for Small-Sample Image Classification}

\author{\IEEEauthorblockN{Xiaoxu Li, Liyun Yu, Xiaochen Yang, Zhanyu Ma, Jing-Hao Xue, Jie Cao, Jun Guo} \thanks{

X. Li, L. Yu and J. Cao are with the School of Computer and Communication, Lanzhou University of Technology, Lanzhou 730050, China.

X. Li, Z. Ma and J. Guo are with the Pattern Recognition and Intelligent System Laboratory, School of Artificial Intelligence, Beijing University of Posts and Telecommunications, Beijing 100876, China.

X. Yang and J.-H. Xue are with the Department of Statistical Science, University College London, London, WC1E 6BT, U.K.}}

\maketitle

\begin{abstract}
Despite achieving state-of-the-art performance, deep learning methods generally require a large amount of labeled data during training and may suffer from overfitting when the sample size is small. To ensure good generalizability of deep networks under small sample sizes, learning discriminative features is crucial. To this end, several loss functions have been proposed to encourage large intra-class compactness and inter-class separability. In this paper, we propose to enhance the discriminative power of features from a new perspective by introducing a novel neural network termed Relation-and-Margin learning Network (ReMarNet). Our method assembles two networks of different backbones so as to learn the features that can perform excellently in both of the aforementioned two classification mechanisms. Specifically, a relation network is used to learn the features that can support classification based on the similarity between a sample and a class prototype; at the meantime, a fully connected network with the cross entropy loss is used for classification via the decision boundary. Experiments on four image datasets demonstrate that our approach is effective in learning discriminative features from a small set of labeled samples and achieves competitive performance against state-of-the-art methods. 
Codes are available at https://github.com/liyunyu08/ReMarNet.
\end{abstract}


\section{Introduction}\label{sec:intro}

Deep learning has achieved state-of-the-art results in various visual tasks, including image and video classification~\cite{karpathy2014large,zhang2019fine,chang2020mutualchannel,wang2018scene,wang2018locality}, object recognition~\cite{wang2019fast,jin2019multi}, and semantic segmentation~\cite{long2015fully}. However, its superior performance heavily relies on a large number of labeled training samples, which are difficult to acquire in many cases, thus severely limiting its application in real life. In addition, when the size of training set is small, the deep model will inevitably suffer from overfitting as the network architecture goes deeper. Hence, how to avoid overfitting and obtain a model with good generalizability under the condition of small sample sizes is a great challenge.

Many methods have been proposed to reduce overfitting in the case of small sample sizes, which can be mainly divided into data enhancement~\cite{antoniou2017data}, domain adaptation~\cite{tzeng2017adversarial,rozantsev2018residual}, regularization~\cite{wan2013regularization,OSLNet2020}, network ensemble~\cite{huang2017snapshot}, and feature extraction~\cite{schroff2015facenet,ma2019insights}. Recently, in the field of feature extraction, there has been a growing number of research on learning discriminative features as a way of preventing overfitting in neural networks. The fundamental pipeline is to optimize a loss function toward better intra-class compactness and inter-class separability. However, most existing methods make assumptions about the type of metric or data distribution beforehand, and these assumptions limit the adaptability of these methods to different tasks.

\begin{figure*}[h]
\centering
\includegraphics[width=0.8\textwidth]{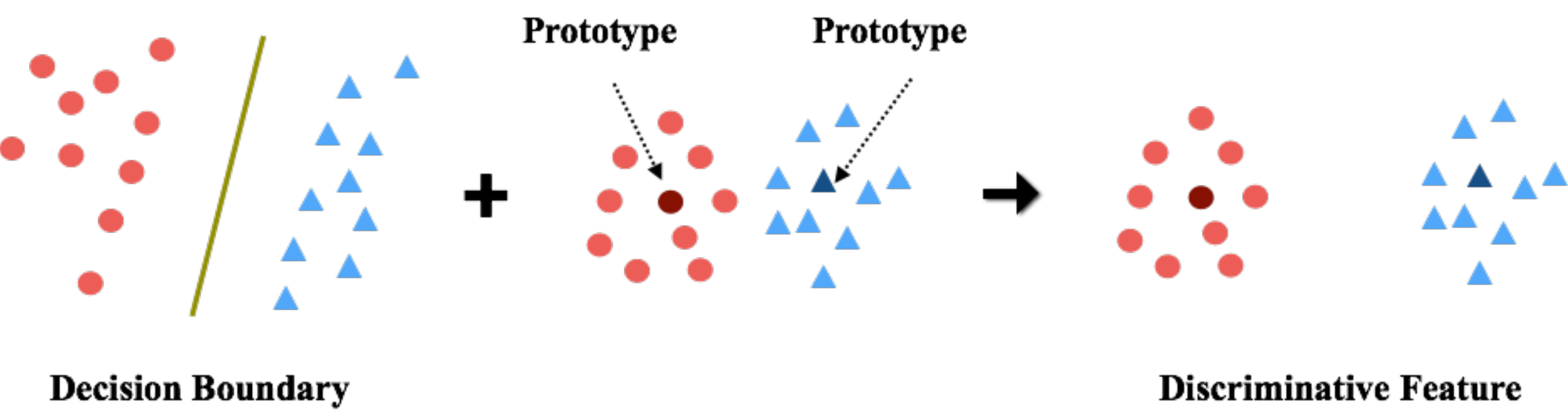}
\caption{The motivation of the proposed Relation-Margin learning neural Network (ReMarNet). Taking an example of binary classification: the round and triangle points represent two classes, respectively. In each class, the sample that an arrow points to denotes the prototype of the class. A green line denotes the decision boundary of two classes.
The discriminability of features will be enhanced if they could excel in both classification paradigms, i.e. the paradigm based on learning decision boundary and the paradigm through comparing the similarity to class prototypes.}\label{motivation}
\end{figure*}

In this paper, we propose a new method for learning the discriminative features and performing classification. Our motivation derives from two aspects. Firstly, as illustrated in Figure~\ref{motivation}, intuitively, if the features can support both the classification paradigm based on learning decision boundary between different classes and the classification paradigm through comparing the similarity to class prototypes, the discriminability of features will be enhanced. Secondly, inspired by the recognition mechanism of human beings, these two classification paradigms are often considered jointly to identify the category of an unseen object; that is, the prediction will be made by considering the outcomes of two paradigms jointly. Building on these two aspects, we propose a \emph{Relation-and-Margin learning neural Network} (ReMarNet) for small-sample image classification, which could perform feature learning and classification from two perspectives jointly.

To implement our proposal, the ReMarNet consists of a feature embedding module and a classification module. The classification module comprises two branches. One branch is constructed from the relation network~\cite{sung2018learning}, which reduces the distance between each sample and its corresponding class prototype and thereby improves the intra-class compactness. The other branch is a two-layer fully connected network with the cross-entropy loss, which guarantees the prediction accuracy. The network is trained in an end-to-end fashion so as to learn the discriminative features that can conjointly learn the satisfactory separation margin and prototype similarity for better small-sample classification. The final prediction is produced by assembling the outputs of two network branches for better generalization. 

To investigate the effectiveness of the proposed method, we conduct experiments on four real image datasets for small-sample classification. Results suggest that assembling two network structures is superior to using a single-branch network and it achieves the state-of-the-art performance compared with existing loss-based methods and two ensemble methods. Our contributions are twofold: 
\begin{itemize}
\item To the best of our knowledge, we propose the first network of integrating two kinds of classification mechanisms, i.e. the classification mechanism based on prototype similarity and the classification mechanism based on decision boundary, termed Relation-and-Margin learning neural network (ReMarNet), for classification with a small number of training samples. It allows for classification separately or conjointly, as preferred by practitioners.

\item Experimental results on four small-sample image datasets show that, compared with the latest work on learning discriminative features via loss functions, our method can obtain more discriminative features and superior performance.
\end{itemize}

\begin{figure*}[htbp]
\begin{center}

\includegraphics[width=0.9\textwidth]{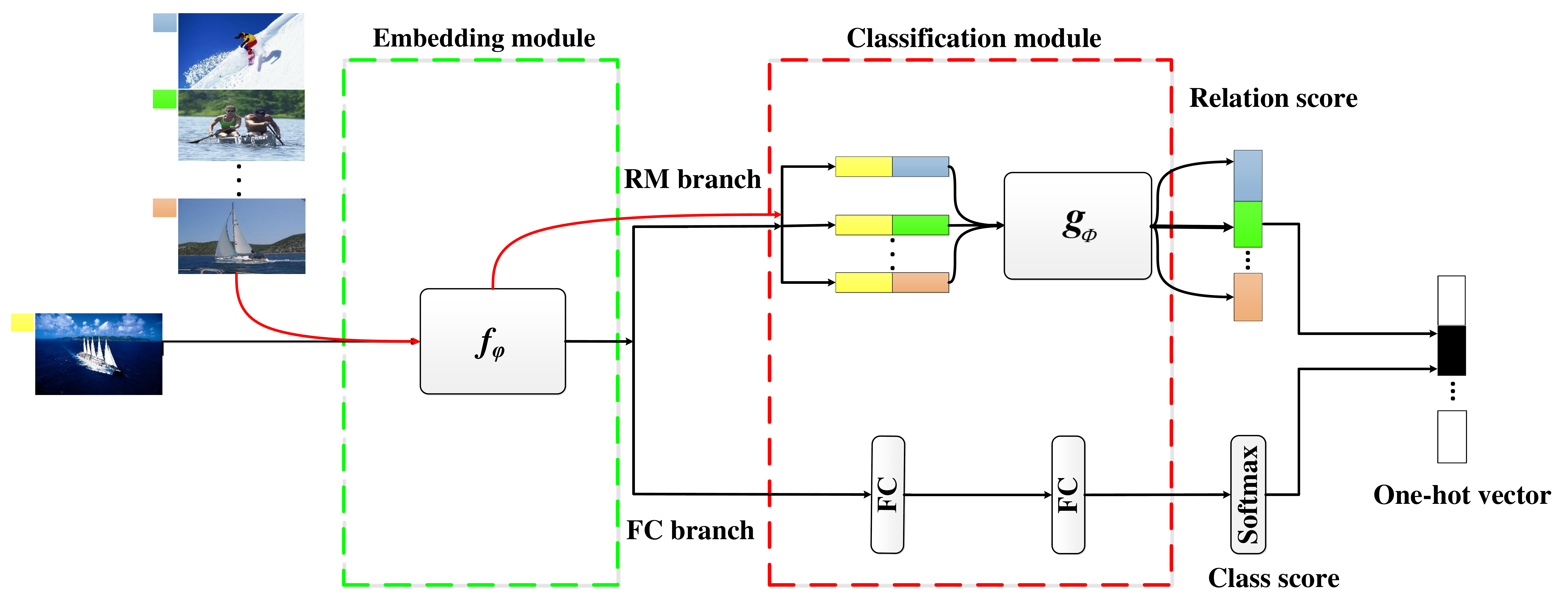}
\end{center} 
\caption{Relation-and-Margin learning neural Network (ReMarNet): The network is composed of a feature embedding module ${f_\varphi}$ and a two-branch classification module, namely the relation module~(RM) branch and the fully connected network~(FC) branch. Images at the top-left of the figure are the prototype samples that we select from each class; each class prototype is assigned with a different color. Given a new sample labeled in yellow, the ReMarNet predicts its class label by leveraging the scores of the two branches. }\label{structure}
 
\end{figure*}

\section{Related Work}
Small-sample learning has received considerable attention in the machine learning field. One category of small-sample learning is few-shot learning. The difference between few-shot learning and the general small-sample learning lies in the evaluation procedure. In few-shot learning~\cite{gidaris2018dynamic,ren2018meta}, the evaluation procedure averages out accuracy over many episodes. Each episode performs a $C$-class classification task, and each class includes $K$ labeled samples; $C$ and $K$ are fixed constants. In the general small-sample learning, the number of classes is determined by the dataset and the number of labeled samples can be unequal. This paper focuses on the general small-sample classification of image data; for few-shot learning, we refer interested readers to~\cite{wang2019few}.

The small sample size poses a challenge to deep learning methods, as they are easy to overfit when the model goes deeper. Data enhancement and domain adaptation methods are proposed to alleviate this problem through increasing the number of training samples. For example, the data-enhanced GAN model can automatically learn to augment training data~\cite{antoniou2017data}. The work in \cite{hariharan2017low} proposes a novel way of transferring the data transformation mode of the base class to generate samples in new categories. Regularization is another widely adopted technique for mitigating overfitting of training networks under small samples. Examples include norm-based constraints \cite{goodfellow2016deep}, dropout~\cite{xiao2016learning,wan2013regularization}, early stopping~\cite{wei2017early}, noise robustness~\cite{surh2017noise}, adversarial training~\cite{peng2018weakly} and multi-task learning~\cite{ren2018cross}. Assembling multiple networks is known to yield more accurate and robust predictions than using a single network. To avoid high computational cost ensued from training multiple networks, Snapshot ensembling~\cite{huang2017snapshot} and temporal ensembling \cite{laine2016temporal} have been proposed, both of which combine multiple outputs obtained from a single training of the network.

Another group of methods focus on learning discriminative features. The pioneering work of~\cite{schroff2015facenet} introduces the triplet loss to separate a positive pair (two matching samples) from a negative one (non-matching samples) by a distance margin in the Euclidean space. Compared with the Euclidean distance, large-margin loss based on the cosine similarity is more appropriate when used in conjunction with the softmax loss, which is widely used in convolutional neural networks~(CNNs) and has demonstrated the capability of learning discriminative features. Building on the link between the cosine similarity and the softmax decision boundary, \cite{liu2016large} and \cite{liu2017sphereface} enforce a stricter condition on the angle between the feature vector and the weight vector so as to improve the discriminative power of the softmax loss. With an $L_2$ normalization of the weight vector, \cite{liu2017sphereface} can be further regarded as imposing a margin in the angular space. Sharing a similar idea, \cite{wang2018cosface} proposes the large margin cosine loss~(LMCL), where the feature vector is additionally normalized and the margin constraint is placed on the cosine similarity, \emph{i.e.}, encouraging a margin in the cosine space. 

Aiming for both intra-class compactness and inter-class separation, the center loss is proposed to punish a large distance between the feature and its corresponding class center, and to jointly supervise the CNNs. It is balanced against the softmax loss via a weight parameter~\cite{wen2016discriminative}. The idea of class centers is adopted in~\cite{liu2017rethinking} but formulated in a different way. Instead of the Euclidean distance, the cosine similarity is calculated and normalized via the softmax function. The loss function is designed in a cross-entropy manner, thus avoiding the weight parameter. \cite{wan2018rethinking} assumes that all the features follow a Gaussian mixture model, and then improves the classification performance by introducing a classification margin and a likelihood regularization term, which includes the center loss as a special case.

\section{ReMarNet: Relation-and-Margin Learning Neural Network}

To learn more discriminative features for small-sample classification, we conjointly enhance the intra-class compactness and enforce the inter-class separability through constructing and simultaneously learning two types of networks. Figure~\ref{structure} summarizes our approach illustratively. After extracting the features via the VGG16 network~\cite{simonyan2014very}, we shrink the distance between all the training samples and their prototype samples via the relation module (RM) to achieve the intra-class compactness, and simultaneously separate the instances from different classes via a two-layer fully connected~(FC) network by the cross-entropy loss to enhance the inter-class separability.
The relation score from the RM and the probability vector from the FC network will be assembled to predict the class label. Before explaining the structure of the proposed ReMarNet in detail, we first review the relation network~\cite{sung2018learning}, on which the RM is built.

\subsection{Relation Network}\label{subsec:RN}

The relation network is proposed in \cite{sung2018learning} for few-shot classification. It is constructed by two modules, namely an embedding module and a relation module. 

The embedding module consists of four convolutional blocks, each of which contains a $3\times 3$ convolution with 64 filters, a batch normalization, and a ReLU nonlinearity layer. In addition, for the first two blocks, a $2\times 2$ max-pooling layer is placed after each block. Two outputs from the embedding module, \emph{i.e.} two feature maps, is concatenated to construct a relation pair, which is used as the input of the subsequent relation module. 

The relation module is composed of two convolutional blocks and two fully connected layers. Each convolution block consists of $3\times 3$ convolution with 64 filters, followed by batch normalization, a ReLU activation function and a $2 \times 2$ max-pooling. The padding parameter of both blocks is set to $1$. The activation functions of all the fully connected layers are ReLU except for the output layer, where a Sigmoid function is adopted in order to generate the relation scores. The input sizes of the first and second FC layers are $64$ and $32$, respectively, and the final output size is $1$.

In summary, the input to the relation module is a concatenated feature map obtained from the embedding module, and the output is a vector of the relation score. The relation module of relation network is adopted in our proposed ReMarNet to learn the similarity between a sample and the prototype of each class.

\subsection{Structure of ReMarNet}
Consider a $K$-class classification task. Let $D_{train}= \{(\bm x_i,\bm y_i)\}_{i=1}^N$ denote a training dataset of $N$ samples, where $\bm y_i$ is a one-hot $K$-dimensional vector representing the class label of $\bm x_i$. For later use in RM, we randomly select one sample from each class of $D_{train}$ as prototype samples and denote them as $\{\bm o_j\}_{j=1}^K$.

The proposed ReMarNet comprises two parts.
The first part is a feature embedding module ($f_\varphi$ in Figure~\ref{structure}). Here we use all the convolutional blocks of the VGG16 network, which produces feature maps \emph{${f_\varphi}$}($\bm x_i$) and \emph{${f_\varphi}$}($\bm o_j$) for samples $\bm x_i$ and $\bm o_j$, respectively. 

The second part is a two-branch classification module, consisting of an RM branch for optimizing the intra-class compactness and an FC branch for pushing the inter-class separability. Let $C(\cdot,\cdot)$ denote the concatenation operator of features maps. The RM branch takes a relation pair of feature maps $f_\varphi(\bm x_i)$ and $f_\varphi(\bm o_j)$, \emph{i.e.} $C(f_\varphi(\bm x_i), f_\varphi(\bm o_j))$, as input, and learns the similarity between them through the network ${g_\phi}$. The output of RM is the relation score $\bm r_{ij}$ between $\bm x_i$ and $\bm o_j$ as 
\begin{equation}
\bm{r}_{ij}=g_\varphi(C(f_\varphi(\bm x_i), f_\varphi(\bm o_j))) \ ,
\end{equation} 
where $\bm r_{ij}$, ranging from zero to one, measures the similarity between the training sample $\bm x_i$ and the class prototype $\bm o_j$. 

To train the RM, we compute the mean square error~(MSE) between the relation score vector $\bm r_i$ and the ground truth label $\bm y_i$. The loss function for the RM branch is
\begin{equation}
L_{RM}  = \dfrac{1}{N}\sum_{i=1}^{N}\sum_{j=1}^{K}{(\bm{r}_{ij}-\bm y_{ij})}^2 \ ,
\end{equation}
where $\bm y_{ij}$, the $j$th element of $\bm y_{i}$, equals one if $\bm x_i$ belongs to the $j$th class and zero otherwise. 
By minimizing the RM loss, we encourage $\bm x_i$ to stay close to its corresponding class prototype, thereby improving the intra-class compactness.

Regarding the FC branch, we use the flattened convolutional features of training samples as input. In the last layer, the softmax activation function is used to calculate the probability $\bm{p}_i$, a $K$-dimensional vector where each element represents the probability that the sample $\bm x_i$ is assigned to each class. The FC network is trained with the following cross entropy~(CE) loss:
\begin{equation}
L_{CE} = - \frac{1}{N}\sum_{i=1}^{N}\bm{y}_i^T\ \log(\bm{p}_i)  \ .
\end{equation}
Minimizing the CE loss promotes learning the features that could increase the probability of assigning $\bm x_i$ to its ground-truth class.

Integrating the RM and FC branches, we obtain the total loss function of the proposed ReMarNet:
\begin{equation}
L=L_{RM} +L_{CE}  \ .
\label{e-loss}
\end{equation} 

In the prediction stage, we calculate the relation score between the test image and each class prototype and the probability of its belonging to each class. The test image is classified as the class with the maximum sum of the two values.

\renewcommand\arraystretch{1.5}
\begin{table*}[htbp]
   \centering
   \caption{Comparison of the proposed ReMarNet with state-of-the-art methods. The mean value (Mean) and standard deviation (Std.) of classification accuracy are reported with the best results in bold.}\label{tab:sota}
    \begin{tabular}{cccccccccc}
    \hline
    \textbf{Datasets} & \textbf{Measure} & Baseline & Center & LGM   & LMCL  & Dual  & Dropout & Snapshot  & \textbf{Ours} \\
    \hline
    \multirow{2}[2]{*}{\textbf{LM}} & \textbf{Mean} & 0.9275 & 0.9219 & 0.9136 & 0.9207 & 0.9298 & 0.9288 & 0.9271 & \textbf{0.9303} \\
          & \textbf{Std.} & 0.0047 & 0.0060 & 0.0075 & 0.0155 & 0.0051 & 0.0045 & 0.0076 & \textbf{0.0067} \\
    \hline
    \multirow{2}[2]{*}{\textbf{UIUC}} & \textbf{Mean} & 0.9476 & 0.9514 & 0.9492 & 0.9492 & 0.9485 & 0.9472 & 0.9437 & \textbf{0.9581} \\
          & \textbf{Std.} & 0.0045 & 0.0032 & 0.0055 & 0.0052 & 0.0040 & 0.0044 & 0.0045 & \textbf{0.0038} \\
    \hline
    \multirow{2}[2]{*}{\textbf{15Scenes}} & \textbf{Mean} & 0.9142 & \textbf{0.9326} & 0.9214 & 0.9243 & 0.9128 & 0.9146 & 0.9143 & 0.9310 \\
          & \textbf{Std.} & 0.0094 & \textbf{0.0037} & 0.0052 & 0.0037 & 0.0052 & 0.0045 & 0.0037 & 0.0025 \\
    \hline
    \multirow{2}[2]{*}{\textbf{BMW}} & \textbf{Mean} & 0.4094 & 0.4274 & 0.2329 & 0.4402 & 0.4363 & 0.4094 & 0.3936 & \textbf{0.4415} \\
          & \textbf{Std.} & 0.0310 & 0.0400  & 0.0478 & 0.0354 & 0.0438 & 0.0356 & 0.0236 & \textbf{0.0364} \\
    \hline
    \end{tabular}
\end{table*}

\begin{figure*}[htbp]
\centering
\begin{minipage}{0.42\linewidth}
\subfigure[The LabelMe Dataset] {\includegraphics[width=2.8in]{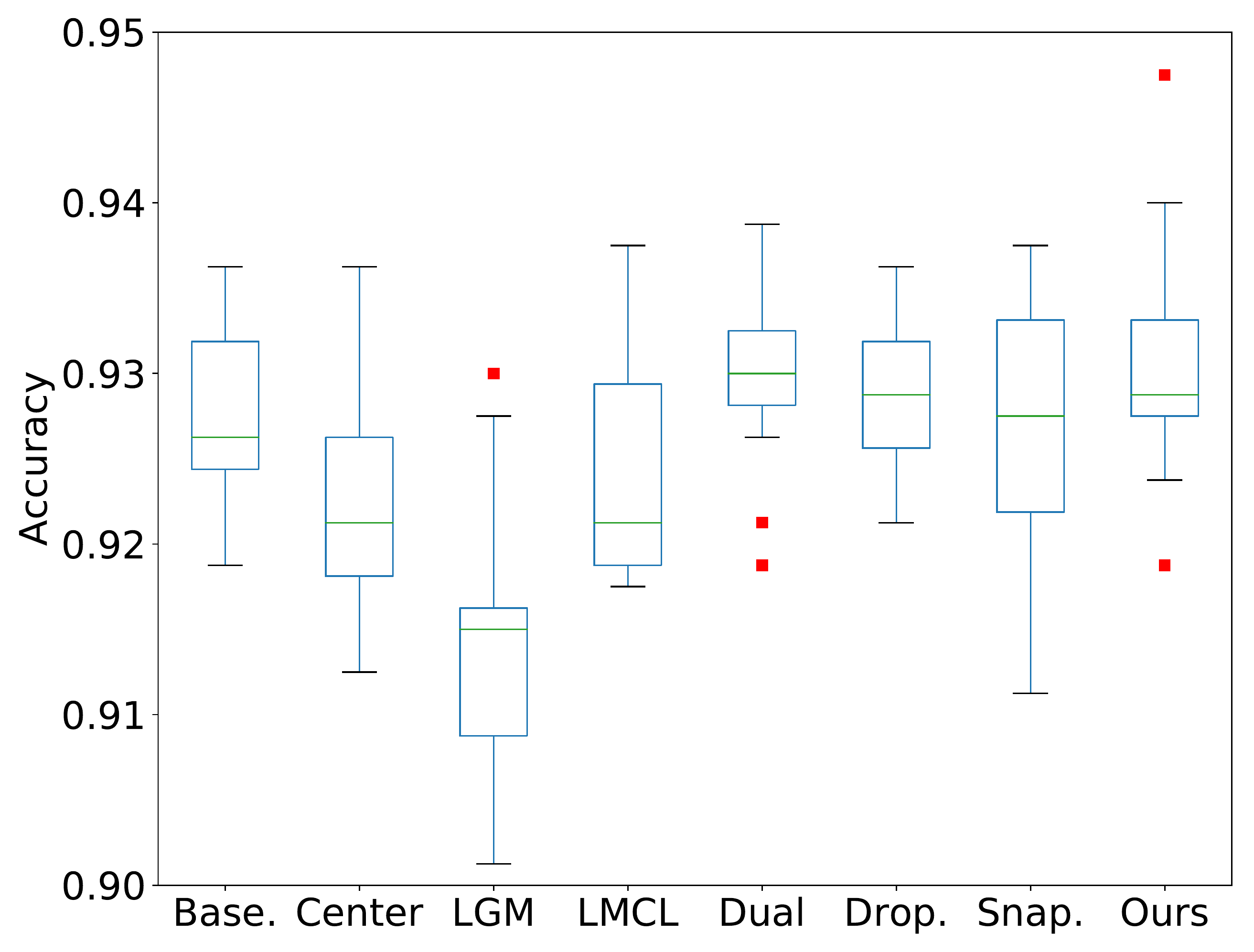}}
\end{minipage} 
\begin{minipage}{0.42\textwidth} 
\subfigure[The UIUC-Sports Dataset] {\includegraphics[width=2.8in]{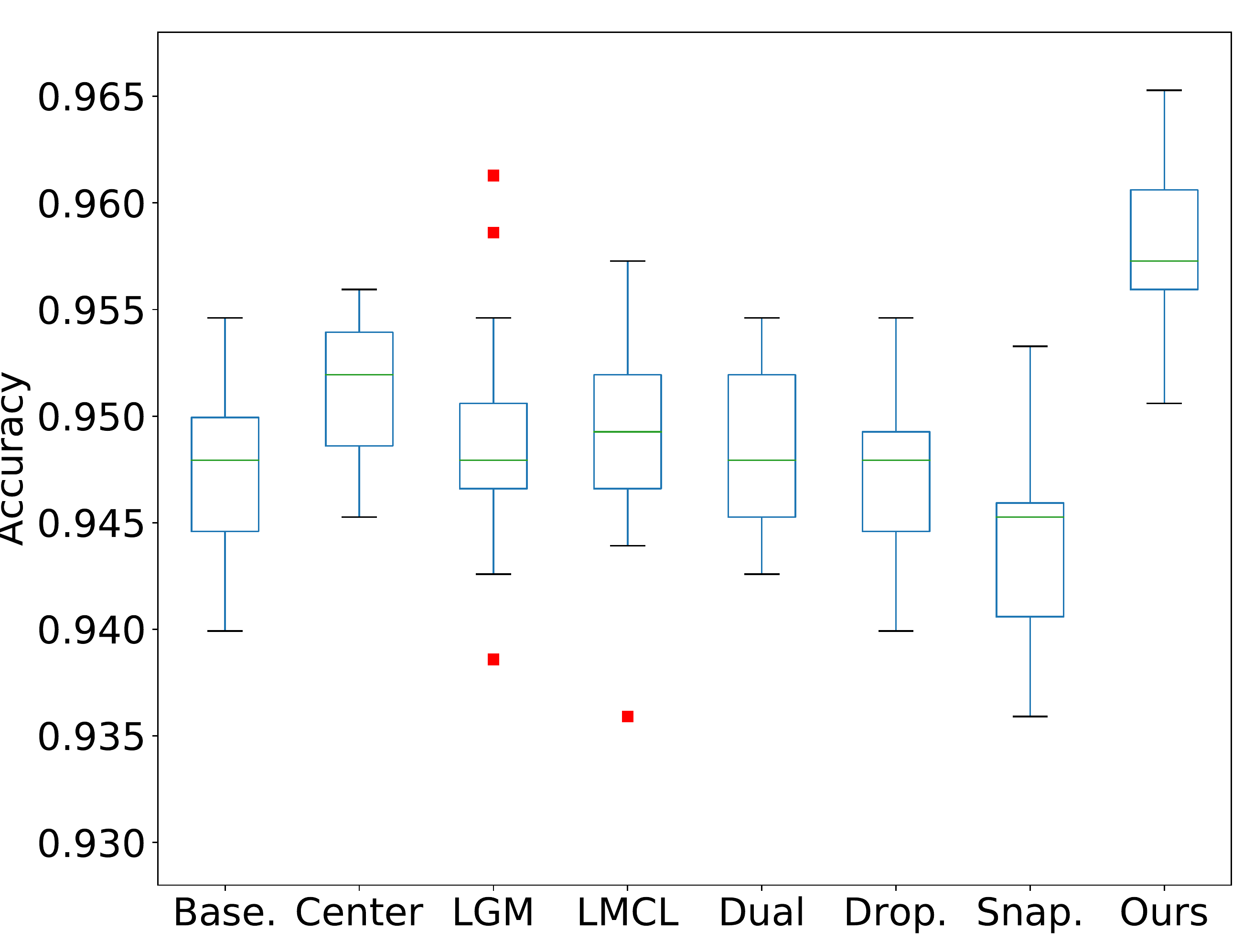}} 
\end{minipage}

\begin{minipage}{0.42\linewidth}
\subfigure[The 15Scenes Dataset] {\includegraphics[width=2.8in]{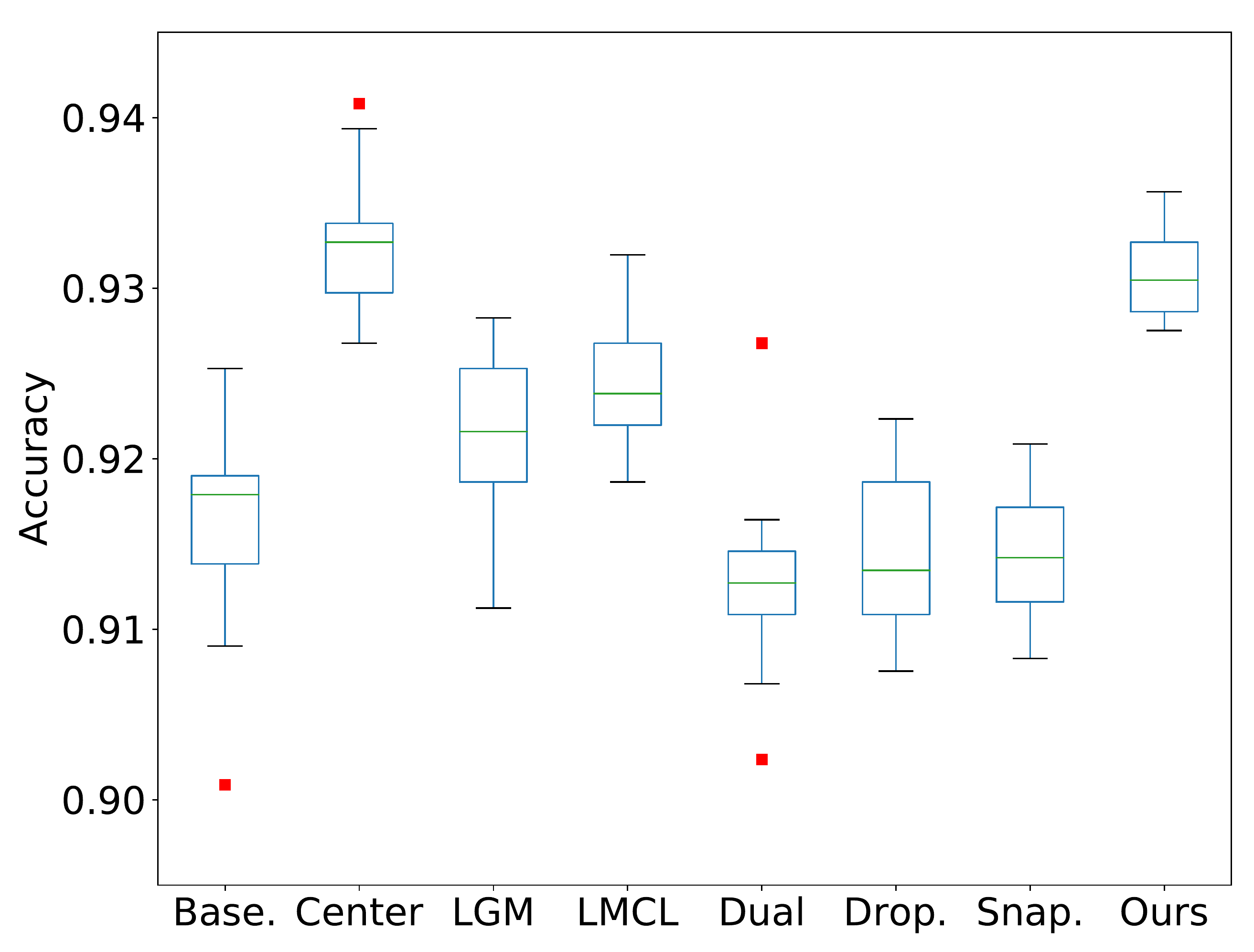}} 
\end{minipage}
\begin{minipage}{0.42\linewidth}
\subfigure[The BMW Dataset] {\includegraphics[width=2.8in]{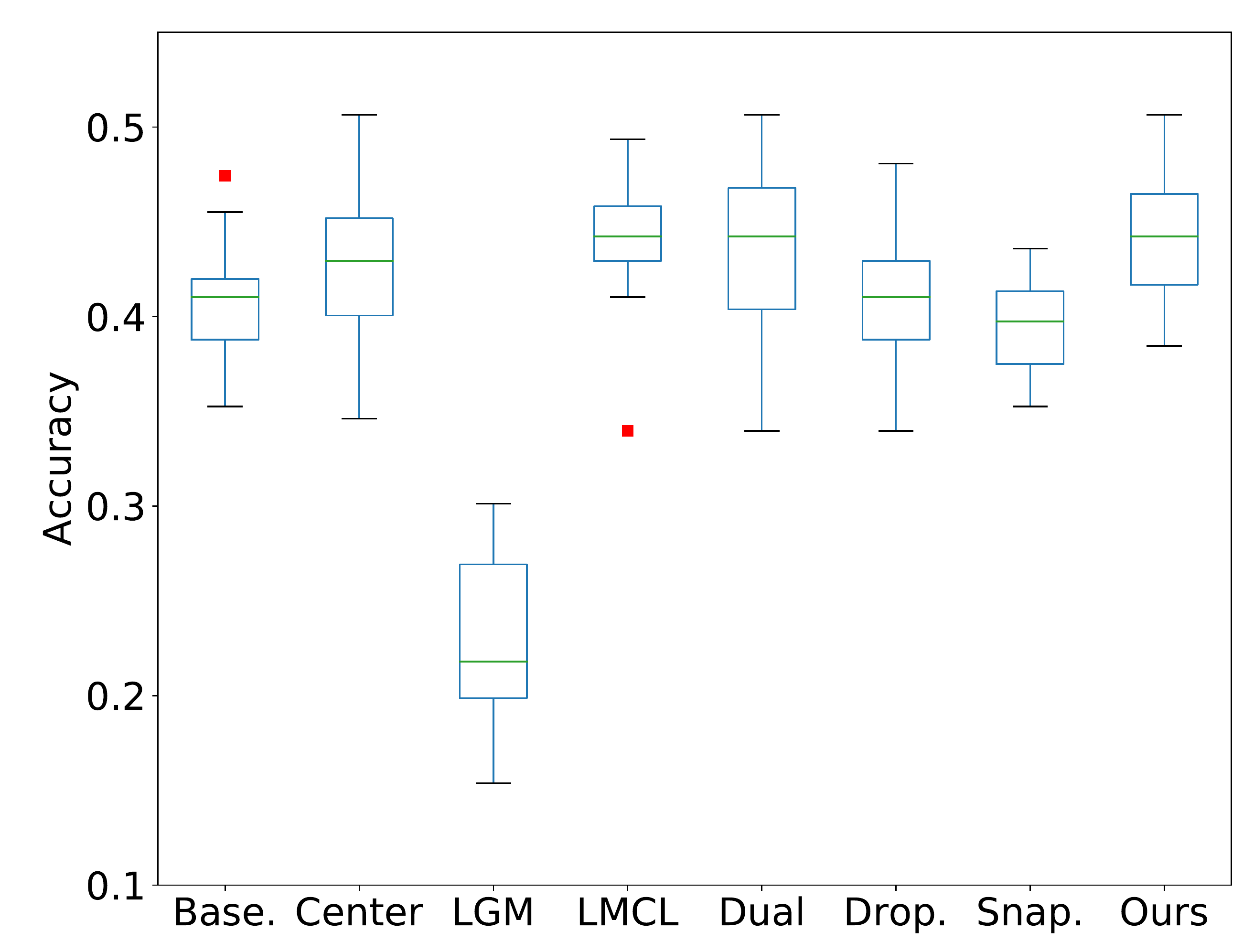}} 
\end{minipage}

\caption{Boxplots of classification accuracy of the proposed ReMarNet and state-of-the-art methods. `Baseline', `Dropout' and `Snapshot' are abbreviated to `Base.', `Drop.', `Snap.', respectively. Each method has been evaluated for $15$ rounds, and the distributions of accuracies are shown via boxplots. In each boxplot, the central mark is the median; the edges of the box are the $25$th and $75$th percentiles, respectively; and the outliers are marked in red individually.
}\label{fig:boxplot}
\end{figure*}
\section{Experimental Results and Analysis}

Experiments in this section serve five purposes: 
\begin{itemize}
    \item To compare the proposed ReMarNet with state-of-the-art methods for the task of small-sample image classification (Sec.~\ref{sec:sota});
    \item To investigate the effect of training set size on different methods (Sec.~\ref{sec:size});
    \item To assess the impact of different backbone networks (Sec.~\ref{sec:Feature-Extractor}); 
    \item To study the effectiveness of each branch of our network (Sec.~\ref{sec:single}, \ref{sec:ensemble}, \ref{sec:Weights});
    \item To evaluate the discriminative power of the learned feature embedding (Sec.~\ref{sec:vis}). 
\end{itemize}

\subsection{Datasets}

For small-sample image classification, we randomly select a subset of images from the following four datasets: LabelMe, UIUC-Sports, 15Scenes and BMW. The datasets vary in their content, number of classes and sample size.

\subsubsection{LabelMe~(LM) Dataset}

LabelMe is a natural scene image classification dataset containing 8 classes: coast, mountain, forest, open country, street, inside city, tall buildings and highways. We randomly select 210 images from each class, of which 100 images are used to form the training set and another 100 images are used for the test set. The total number of images used in each round is 1600. 

\subsubsection{UIUC-Sports Dataset}

UIUC-Sports contains 1578 sports scene images of 8 classes: bocce (137), polo (182), rowing (250), sailing (190), snowboarding (190), rock climbing (194), croquet (236) and badminton (200). A training set of 749 images and a test set of 749 images are randomly sampled from the entire dataset. 

\subsubsection{15Scenes Dataset}

15Scenes is one of the most complete datasets for scene classification used to date in the literature, gradually built from eight classes~\cite{oliva2001modeling} to 13 classes~\cite{fei2005bayesian} and finally to 15 classes~\cite{lazebnik2006beyond}. The total number of images is 4485 and the number per category varies between 200 and 400. The dataset is partitioned into 70\% for training and 30\% for testing. 

\subsubsection{BMW Dataset}
BMW-10 is an ultra-small, fine-grained vehicle dataset comprised of 10 different types of BMW vehicles~\cite{krause20133d}. It contains a total of 512 images, and each class has around 50 images. The training and test ratio is set as 70/30. 

For all datasets, we run 15 rounds of random training and test split. The mean value and standard deviation of classification accuracy are used as our evaluation criteria.

\subsection{Methodologies and Parameter Settings}

We compare the proposed ReMarNet with the baseline method, i.e., ~a fully connected network using the cross entropy loss (Baseline), and four state-of-the-art feature learning methods using different loss functions, namely center loss (Center)~\cite{wen2016discriminative}, L-GM loss (LGM)~\cite{wan2018rethinking}, LMCL loss (LMCL)~\cite{wang2018cosface}, and dual loss (Dual)~\cite{li2019dual}. We also consider two ensembling networks, namely Dropout~\cite{srivastava2014dropout} and Snapshot~\cite{huang2017snapshot}. 

In the proposed ReMarNet, we use the VGG16 network as our feature extractor. The number of hidden layers in the FC branch is set to 32 and details of the RM branch are explained in Sec.~\ref{subsec:RN}.

All networks are trained by using the RMSprop optimizer~\cite{tieleman2012lecture} with a batch size of 32. Learning rates of 0.00001 and 0.0001 are used for the feature extraction network and the FC network, respectively. For Center, LGM and LMCL, we use the stochastic gradient descent algorithm to separately train the loss function with a learning rate of 0.01. For our method, the learning rate of the RM branch is set as 0.001. The number of epochs is 50 for all methods except for the Snapshot network where two models are trained with 50 epochs each, leading to a total of 100 epochs. LGM, LMCL and Dual involve some additional hyperparameters, which are chosen as follows: the loss weight and $\alpha$ in LGM are set as 0.001 and 1.5 respectively; $s$ and $m$ in LCML are set as the average of $\|\bm x\|_2$ and $0.5$ respectively; the loss weight in Dual is set as 4.5.

\subsection{Comparison with State-of-the-art Methods}\label{sec:sota}

Table~\ref{tab:sota} shows the mean value and standard deviation of classification accuracy over 15 rounds of experiments and Figure~\ref{fig:boxplot} depicts the boxplots of the classification accuracy.

As shown in Table~\ref{tab:sota}, all the existing methods cannot consistently outperform the baseline. Methods that target at learning discriminative features, \emph{i.e.}~Center, LGM and LMCL, are inferior to the baseline on the LM dataset, which may indicate that they sacrifice classification accuracy for small intra-class distance or large inter-class margin and underfit the data. Moreover, LGM performs poorly on the BMW dataset, which may be due to the mismatch between the distributional assumption of LGM and the real data. The dual loss is proposed to alleviate the vanishing gradient problem from using the cross entropy loss. Therefore, unless the problem occurs, we would expect its performance to be similar to the baseline. Such a pattern is found in our experiments, where only on the BMW dataset Dual improves the baseline by a large amount. Regarding the ensemble methods, we observe that the performance of Dropout is almost identical to that of the baseline and Snapshot performs worse than the baseline on three datasets. A potential justification for the unsatisfactory performance of Snapshot is as follows. For fairness of comparison, the learning rate is set to be the same for all methods and this value may be too small for Snapshot to reach local minima. Larger learning rates have been tested and we observe a degradation in the performance of Baseline. The proposed ReMarNet always outperforms the baseline; compared with state-of-the-art methods, it also achieves the highest accuracy on three datasets. 

\renewcommand\arraystretch{1.3}
\begin{table}[t]
  \centering
  \caption{$p$-values of the Wilcoxon signed-rank test. $\ast$ indicates that ReMarNet is significantly different from the compared method at the significance level of 5\%. }
    \setlength{\tabcolsep}{0.3mm}{
    \begin{tabular}{cccccccc}
    \hline
    \textbf{Datasets} & \textbf{Baseline} & \textbf{Center} & \textbf{LGM} & \textbf{LMCL} & \textbf{Dual} & \textbf{Dropout} & \textbf{Snapshot } \\
    \hline
    \textbf{LM} & 0.1228  & 0.0006$^\ast$  & 0.0015$^\ast$  & 0.0153$^\ast$  & 0.8438  & 0.6374  & 0.3107  \\
    \hline
    \textbf{UIUC} & 0.0007$^\ast$  & 0.0014$^\ast$  & 0.0021$^\ast$  & 0.0007$^\ast$  & 0.0006$^\ast$  & 0.0007$^\ast$  & 0.0010$^\ast$  \\
    \hline
    \textbf{15Scenes} & 0.0007$^\ast$  & 0.1635  & 0.0007$^\ast$  & 0.0010$^\ast$  & 0.0007$^\ast$  & 0.0007$^\ast$  & 0.0007$^\ast$  \\
    \hline
    \textbf{BMW} & 0.0355$^\ast$  & 0.6374  & 0.0005$^\ast$  & 0.9773  & 0.6694  & 0.0008$^\ast$  & 0.0230$^\ast$  \\
    \hline
    \end{tabular}}%
  \label{tab:W-test}%
\end{table}%
To further demonstrate the superiority of the proposed method, we conduct the Wilcoxon signed-rank tests \cite{wilcoxon1945individual} between the ReMarNet and other referred methods. The Wilcoxon signed-rank test is a non-parametric statistical hypothesis test, used to check the existence of significant difference between each pair of methods. Table~\ref{tab:W-test} suggests that, at the significance level of 5\%, ReMarNet is significantly different from others in the majority of cases.

We now focus on the reliability of the proposed method. Figure~\ref{fig:boxplot}  shows boxplots of classification accuracy. Again, we observe that the existing methods outperform the baseline on some datasets only and deteriorate on at least one dataset, whereas our proposed ReMarNet achieves higher median accuracy on all four datasets and maintains similar spread as indicated by the interquartile range~(IQR). On the LabelMe and BMW datasets, our method obtains higher first, second and third quartiles compared with the baseline. Its advantage is more pronounced on the UIUC and 15Scenes datasets. On the UIUC dataset, all methods have similar IQR but our method has a much higher median value. On the 15Scenes dataset, the proposed method has a smaller IQR than Baseline and the worst performance of ours is still larger than the baseline. Except on the LabelMe dataset, our method does not produce any outlier. 

Another aspect of reliability is the method's stability to random choices of prototypes. The class prototypes are currently selected in a random manner at the beginning of the network training, and are fixed until the evaluation procedure finishes. To monitor the performance change of ReMarNet on the LM dataset, we sample $9$ different sets of prototype images and run the ReMarNet for 15 rounds on each set of prototype images. Among $9$ sets of experiments, the  highest mean accuracy is $0.9339$ and the lowest one is $0.9303$; the standard deviation is $0.0067$ in both cases. This result shows that the proposed ReMarNet remains stable as the class prototypes change.

\renewcommand\arraystretch{1.5}
\begin{table*}[htbp]
  \centering
  \caption{
  Comparison of classification accuracy under different training sample sizes. The notation `DatasetName-$n$' denotes that the number of training samples per class is reduced by $n$ from its original size.}\label{tab:size}
    \begin{tabular}{cccccccccc}
    \hline
    \textbf{Datasets} & \textbf{Measure} & Baseline & Center & LGM & LMCL & Dual & Dropout & Snapshot & \textbf{Ours} \\
    \hline
    \multirow{2}[2]{*}{\textbf{LM-20}} & \textbf{Mean}  & 0.9248 & 0.9116 & 0.9015 & 0.9065 & 0.9252 & 0.9247 & 0.9168 & \textbf{0.9262} \\
          & \textbf{Std.}   & 0.0059 & 0.0089 & 0.0095 & 0.0081 & 0.0045 & 0.0062 & 0.0101 & \textbf{0.0054} \\
    \hline
    \multirow{2}[2]{*}{\textbf{LM-40}} & \textbf{Mean}  & 0.9148 & 0.9180 & 0.9026 & 0.9138 & 0.9151 & 0.9148 & 0.9123 & \textbf{0.9215} \\
          & \textbf{Std.}   & 0.0055 & 0.0079 & 0.0087 & 0.0090 & 0.0062 & 0.0048 & 0.0059 & \textbf{0.0064} \\
    \hline
    \multirow{2}[2]{*}{\textbf{LM-60}} & \textbf{Mean}  & 0.9035 & 0.8947 & 0.8813 & 0.8971 & 0.9064 & 0.9036 & 0.9028 & \textbf{0.9082} \\
          & \textbf{Std.}   & 0.0056 & 0.0085 & 0.0126 & 0.0077 & 0.0067 & 0.0053 & 0.0085 & \textbf{0.0081} \\
    \hline
    \multirow{2}[2]{*}{\textbf{LM-80}} & \textbf{Mean}  & 0.8928 & 0.8933 & 0.8896 & 0.9011 & 0.8913 & 0.8917 & 0.8870 & \textbf{0.9015} \\
          & \textbf{Std.}   & 0.0098 & 0.0081 & 0.0081 & 0.0116 & 0.0119 & 0.0100  & 0.0170 & \textbf{0.0084} \\
    \hline
    \multirow{2}[2]{*}{\textbf{UIUC-10}} & \textbf{Mean}  & 0.9438 & 0.9531 & 0.9475 & 0.9463 & 0.9447 & 0.9443 & 0.9426 & \textbf{0.9566} \\
          & \textbf{Std.}   & 0.0041 & 0.0032 & 0.0062 & 0.0042 & 0.0064 & 0.0043 & 0.0050 & \textbf{0.0038} \\
    \hline
    \multirow{2}[2]{*}{\textbf{UIUC-20}} & \textbf{Mean}  & 0.9421 & 0.9456 & 0.9414 & 0.9429 & 0.9401 & 0.9401 & 0.9422 & \textbf{0.9510} \\
          & \textbf{Std.}   & 0.0040 & 0.0035 & 0.0081 & 0.0045 & 0.0051 & 0.0048 & 0.0060 & \textbf{0.0042} \\
    \hline
    \multirow{2}[2]{*}{\textbf{UIUC-30}} & \textbf{Mean}  & 0.9379 & 0.9340 & 0.9364 & 0.9366 & 0.9344 & 0.9372 & 0.9336 & \textbf{0.9448} \\
          & \textbf{Std.}   & 0.0048 & 0.0037 & 0.0054 & 0.0059 & 0.0046 & 0.0052 & 0.0060 & \textbf{0.0053} \\
    \hline
    \multirow{2}[2]{*}{\textbf{UIUC-40}} & \textbf{Mean}  & 0.9211 & 0.9251 & 0.9263 & 0.9266 & 0.9203 & 0.9213 & 0.9211 & \textbf{0.9301} \\
          & \textbf{Std.}   & 0.0081 & 0.0046 & 0.0064 & 0.0073 & 0.0061 & 0.0084 & 0.0073 & \textbf{0.0060} \\
    \hline
    \end{tabular}
\end{table*}

\renewcommand\arraystretch{1.7}

\begin{table*}[htbp]
 \centering
  \caption{Comparison of classification accuracy on the LM dataset with different networks as the feature extractor.}
    \begin{tabular}{cccccccccc}
    \hline
    \textbf{Network} & \textbf{Measure} & Baseline & Center & LGM & LMCL & Dual & Dropout & Snapshot & \textbf{Ours} \\
    \hline
    \multirow{2}[1]{*}{\textbf{VGG16}} & \textbf{Mean} & 0.9275 & 0.9219 & 0.9136 & 0.9207 & 0.9298 & 0.9288 & 0.9271 & \textbf{0.9303} \\
    & \textbf{Std.} & 0.0047 & 0.006 & 0.0075 & 0.0155 & 0.0051 & 0.0045 & 0.0076 & \textbf{0.0067} \\
\hline
\multirow{2}[2]{*}{\textbf{AlexNet}} & \textbf{Mean} & 0.8982 & 0.9013 & 0.8844 & 0.9006 & 0.8958 & 0.8976 & 0.9006 & \textbf{0.9103} \\
          & \textbf{Std.} & 0.0051 & 0.0071 & 0.0172 & 0.0085 & 0.0058 & 0.0053 & 0.0058 & \textbf{0.0050} \\
\hline
\multirow{2}[2]{*}{\textbf{DenseNet-121}} & \textbf{Mean} & 0.8846 & 0.8897 & $-$   & 0.8880 & 0.8901 & 0.8846 & 0.8895 & \textbf{0.8937} \\
          & \textbf{Std.} & 0.0109 & 0.0068 & $-$    & 0.0090 & 0.0088 & 0.0100  & 0.0105 & \textbf{0.0091} \\
    \hline
    \end{tabular}%
  \label{tab:net}%
\end{table*}%

\subsection{Performance Evaluation under Different Training Sizes}\label{sec:size}

To further evaluate classification performance of all methods in the small-sample size setting, we reduce the training sample size in the LabelMe and UIUC-Sports datasets. For LableMe, the number of training samples per class is reduced by 20, 40, 60 and 80 from its original size; the reduced datasets are denoted as LM-20, LM-40, LM-60, LM-80, respectively. For UIUC-Sports, we reduce 10, 20, 30, and 40 samples from each class and datasets are denoted in a similar way. The number of samples in the validation and test sets remains unchanged. The mean value and standard deviation of classification accuracy are listed in Table~\ref{tab:size}.

As the training set gets smaller, it becomes more difficult to learn discriminative features. As anticipated, the accuracy of each method decreases with the number of reduced samples. On the LM dataset, Dual and Dropout, which originally outperform Baseline without data reduction, lose their advantages when the number of training samples is reduced by 80. Similar observations are found on the UIUC dataset, where Center, LGM, LMCL and Dual all perform worse than Baseline when reducing the training size by 30. In contrast, the proposed ReMarNet maintains its superiority over all methods across different sample sizes. 

\begin{figure*}[htbp]
\begin{center}
\begin{minipage}{0.43\linewidth}
\subfigure[The LabelMe Dataset] {\includegraphics[width=3.1in]{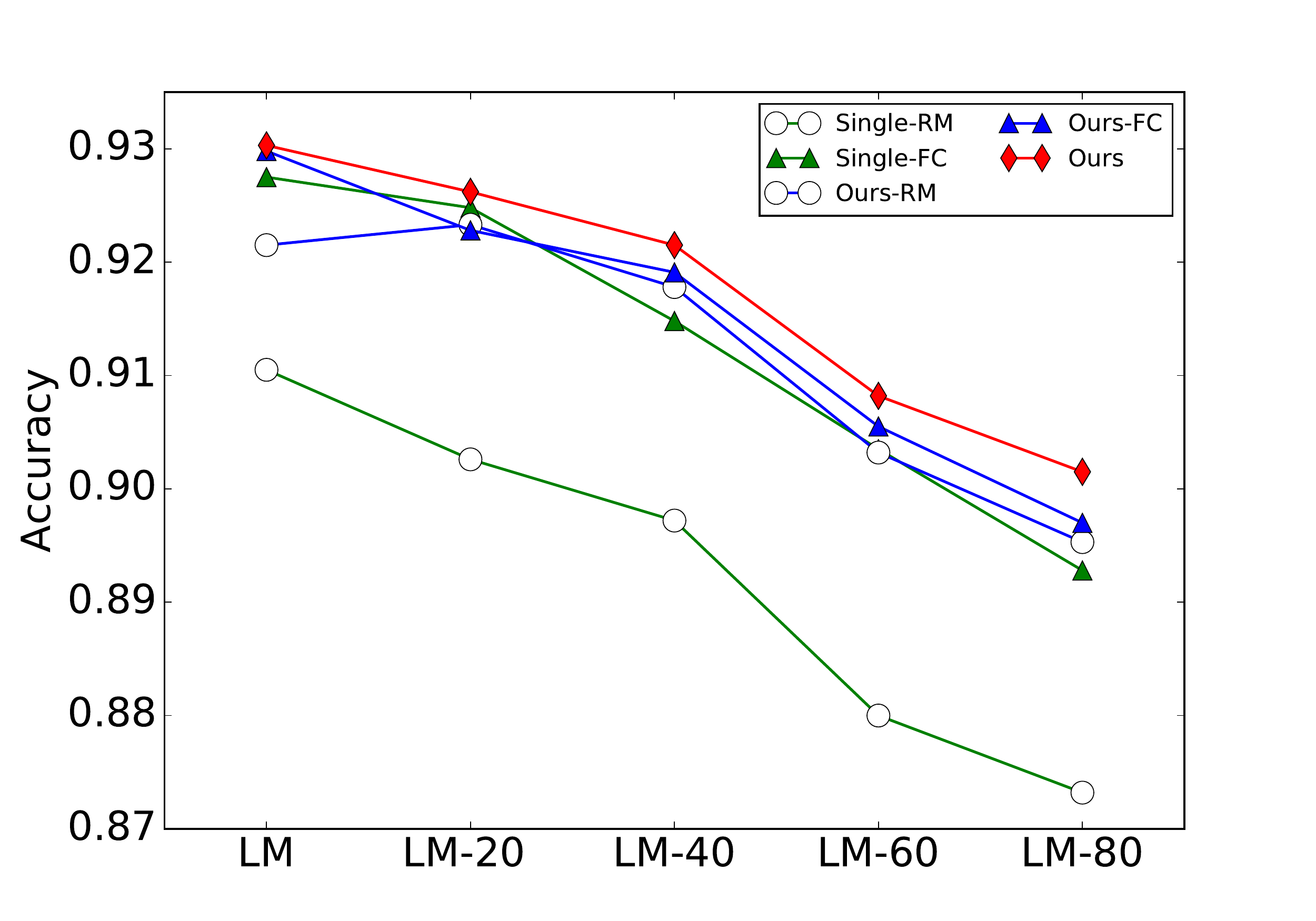}}
\end{minipage} 
\begin{minipage}{0.43\textwidth} 
\subfigure[The UIUC-Sports Dataset] {\includegraphics[width=3.1in]{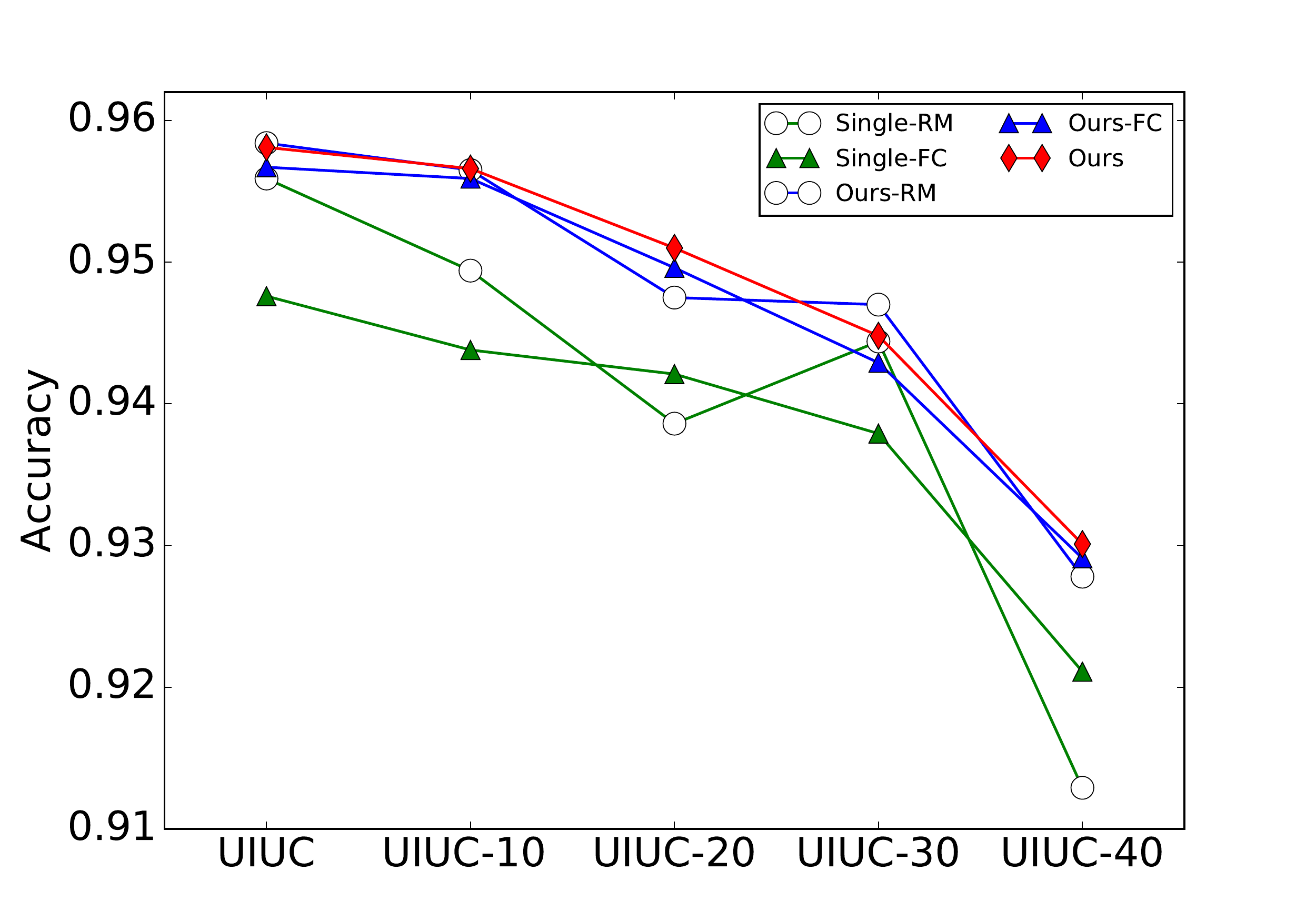}} 
\end{minipage}
\end{center} 
\caption{
Comparison of classification accuracy obtained by Single-RM, Single-FC, RM Branch (Ours-RM) and FC Branch (Ours-FC) in our method, as well as the proposed ReMarNet (Ours). In Single-RM and Single-FC, training and prediction are based on RM or FC only; in Ours-RM and Ours-FC, training is based on both RM and FC and prediction is based on RM or FC only; in Ours, training and prediction are based on both RM and FC.}\label{fig:ensemble}
\end{figure*}

\renewcommand\arraystretch{1.3}
\begin{table}[htbp]
  \begin{center}
  \caption{Comparison of classification accuracy obtained from single-branch Relation Network (Base.-RM), single-branch fully connected network (Base.-FC) and the proposed ReMarNet (Ours). `DatasetName-$n$' denotes that the number of training samples per class is reduced by $n$.}\label{tab:single}%
  \setlength{\tabcolsep}{3.7mm}{
    \begin{tabular}{ccccc}
    \hline
    \textbf{Dataset}  & \textbf{Measure} & Base.-RM & Base.-FC  & \textbf{Ours} \\
    \hline
    \multirow{2}[2]{*}{\textbf{LM}} & \textbf{Mean} & 0.9105 & 0.9275 & \textbf{0.9303} \\
          & \textbf{Std.} & 0.0088 & 0.0047 & \textbf{0.0067} \\
    \hline
    \multirow{2}[2]{*}{\textbf{LM-20}} & \textbf{Mean} & 0.9026 & 0.9248 & \textbf{0.9262} \\
          & \textbf{Std.} & 0.0119 & 0.0059 & \textbf{0.0054} \\
    \hline
    \multirow{2}[2]{*}{\textbf{LM-40}} & \textbf{Mean} & 0.8972 & 0.9148 & \textbf{0.9215} \\
          & \textbf{Std.} & 0.0079 & 0.0055 & \textbf{0.0064} \\
    \hline
    \multirow{2}[2]{*}{\textbf{LM-60}} & \textbf{Mean} & 0.8800  & 0.9035 & \textbf{0.9082} \\
          & \textbf{Std.} & 0.0120 & 0.0056 & \textbf{0.0081} \\
    \hline
    \multirow{2}[2]{*}{\textbf{LM-80}} & \textbf{Mean} & 0.8732 & 0.8928 & \textbf{0.9015} \\
          & \textbf{Std.} & 0.0135 & 0.0098 & \textbf{0.0084} \\
    \hline
    \multirow{2}[2]{*}{\textbf{UIUC}} & \textbf{Mean} & 0.9559 & 0.9476 & \textbf{0.9581} \\
          & \textbf{Std.} & 0.0055 & 0.0045 & \textbf{0.0038} \\
    \hline
    \multirow{2}[2]{*}{\textbf{UIUC-10}} & \textbf{Mean} & 0.9494 & 0.9438 & \textbf{0.9566} \\
          & \textbf{Std.} & 0.0051 & 0.0041 & \textbf{0.0038} \\
    \hline
    \multirow{2}[2]{*}{\textbf{UIUC-20}} & \textbf{Mean} & 0.9386 & 0.9421 & \textbf{0.9510} \\
          & \textbf{Std.} & 0.0077 & 0.0040 & \textbf{0.0042} \\
    \hline
    \multirow{2}[2]{*}{\textbf{UIUC-30}} & \textbf{Mean} & 0.9444 & 0.9379 & \textbf{0.9448} \\
          & \textbf{Std.} & 0.0087 & 0.0048 & \textbf{0.0053} \\
    \hline
    \multirow{2}[2]{*}{\textbf{UIUC-40}} & \textbf{Mean} & 0.9129 & 0.9211 & \textbf{0.9301} \\
          & \textbf{Std.} & 0.0081 & 0.0081 & \textbf{0.0060} \\
    \hline
    \end{tabular}}%
    \end{center}
\end{table}%

\subsection{Ablation Study on the Impact of Different Backbone Networks}\label{sec:Feature-Extractor}
In the above experiments, ReMarNet and other compared methods adopt VGG16 as the backbone network. To further explore the potential of the proposed method, we use AlexNet~\cite{krizhevsky2012imagenet} and DenseNet-121~\cite{huang2017densely} to construct ReMarNet and other compared methods, and run the experiment for 15 rounds on the LM dataset. The classification results are listed in Table~\ref{tab:net}. The performance of LGM is not listed when DenseNet-121 is used as the feature extractor as the method cannot fit the training data within 50 epochs.

From the table, we observe that all methods based on VGG16 perform better than their counterparts based on AlexNet and DenseNet-121; our method again outperforms all compared methods. This result shows that the proposed ReMarNet is effective even when the backbone network is changed.

\subsection{Ablation Study on the Effectiveness of Two-branch Network}\label{sec:single}

We now examine the structure of the proposed ReMarNet. Specifically, we compare the results of using only the relation module branch in the classification module (Base.-RM), using only the fully connected network branch (Base.-FC), and the proposed ReMarNet (Ours) which can be regarded as an ensemble of the RM and the FC network. The experimental results are shown in Table~\ref{tab:single}.

First, we notice that the classification accuracy declines more rapidly in Base.-RM than Base.-FC. The reason might be that Base.-RM is more sensitive to the decrease in the number of pairs of feature embedding than Base.-FC. Second, combining these two networks, i.e. using the proposed ReMarNet, certainly provides a performance boost. The reason is that the RM branch and the FC branch have different classification paradigms, and put different assumptions during the training of network. As the ReMarNet trains the two branches conjointly, it will force the network to consider two kinds of assumptions in training and test procedures. Therefore, the generalization performance is enhanced in ReMarNet.

\subsection{Ablation Study on the Effectiveness of Simultaneous Training and Prediction of Two-branch Network}\label{sec:ensemble}

As mentioned in Sec.~\ref{sec:intro}, our motivations behind the two-branch network are that the features allowing for two paradigms of classification should be more discriminative and that the decisions building on two paradigms should be more reliable. To verify if the proposed ReMarNet could achieve the above two goals, we design this experiment to evaluate the performance of the relation module branch trained on its own (Single-RM), the fully connected network branch trained on its own (Singe-FC), RM trained jointly with FC (Ours-RM), FC trained jointly with RM (Ours-FC), and the proposed ReMarNet (Ours). For Single-RM (Single-FC, resp.), we train the classification module with RM (FC network, resp.) only and the class label is predicted based on the relation score (probability vector, resp.); for Ours-RM (Ours-FC, resp.), we train the classification module with both RM and the FC network by using the proposed ReMarNet and then make predictions separately from the relation score (probability vector, resp.); for Ours, RM and the FC network are trained simultaneously and the prediction is made by summing up the two outputs. Figure~\ref{fig:ensemble} shows mean accuracy on the LabelMe and UIUC-Sports datasets.

By comparing single RM and Ours-RM, it is clearly evident that simultaneous training of two networks could generate more discriminative features, and, consequently, improve classification of each network. Similar observation can be found for the FC network, in particular on the UIUC dataset. By comparing Ours against Ours-RM and Ours-FC, we can see that the proposed ensembling network further enhances classification on the LabelMe dataset and the performance gain is more significant when the training set gets smaller. On the UIUC dataset, such improvement can still be observed in most cases. These encouraging results validate our motivation for simultaneous training and prediction on two networks.  

\renewcommand\arraystretch{1.3}
\begin{table}[t]
\centering
  \caption{Classification accuracy of ReMarNet with different weights assigned to the two-branch losses in Eq.~(\ref{e-loss}). $a$ denotes the weight of the RM loss and $b$ denotes the weight of the FC loss.}
  \setlength{\tabcolsep}{3.7mm}{
    \begin{tabular}{c|c|c|c|c}
    \hline
     \multicolumn{1}{c|}{\textbf{Dataset}} & \multicolumn{1}{c|}{$a$} & \multicolumn{1}{c|}{$b$} & \multicolumn{1}{c|}{\textbf{Mean}} & \textbf{Std.} \\
    \hline
    \multirow{6}[24]{*}{\textbf{LM}} & \textbf{0} & \multirow{3}[12]{*}{\textbf{1}} & 0.9306 & 0.0053 \\
          & \textbf{0.2} &       & 0.9305 & 0.0053 \\
          & \textbf{0.4} &       & 0.9332 & 0.0058 \\
          & \textbf{0.6} &       & 0.9322 & 0.0052 \\
          & \textbf{0.8} &       & 0.9322 & 0.0050 \\
          & \textbf{1 } &       & 0.9303 & 0.0067 \\
\cline{2-5}          & \multirow{3}[12]{*}{\textbf{1}} & \textbf{0} & 0.9160 & 0.0046 \\
          &       & \textbf{0.2} & 0.9345 & 0.0043 \\
          &       & \textbf{0.4} & 0.9337 & 0.0061 \\
          &       & \textbf{0.6} & 0.9325 & 0.0067 \\
          &       & \textbf{0.8} & 0.9334 & 0.0050 \\
          &       & \textbf{1} & 0.9303 & 0.0067 \\
    \hline
    \end{tabular}}%
  \label{tab:weight}%
\end{table}%

\subsection{Performance Evaluation with Different Weights of Losses}\label{sec:Weights}
In all the above experiments, our method is optimized by minimizing the total loss as presented in Eq.~(\ref{e-loss}) for simplicity. A more deliberate choice is to compute the weighted sum of RM loss and FC loss. In this section, we change the weight of these two losses and monitor the performance of our method. Specifically, let $a$ and $b$ denote the coefficient of the RM loss and the FC loss, respectively; the previous setting corresponds to $a=1$ and $b=1$. We now fix $a=1$ and vary $b$ from $0$ to $1$, and likewise fix $b=1$ and vary $a$ from $0$ to $1$. The classification performance of our method on the LM dataset is listed in Table~\ref{tab:weight}.

The accuracy remains competitive unless $b=0$ where the FC network is no longer fine-tuned, suggesting the simultaneous training of two networks. Moreover, the classification performance can be improved by varying $a$ or $b$. However, given the performance gain is relatively small and our task of interest is the small-sample classification, it would be preferable to avoid the introduction of the weight parameter and keep the model selection simple.

\subsection{Feature Visualization}\label{sec:vis}

\begin{figure*}[htbp]
\begin{minipage}{0.31\textwidth} 
\includegraphics[width=2.5in]{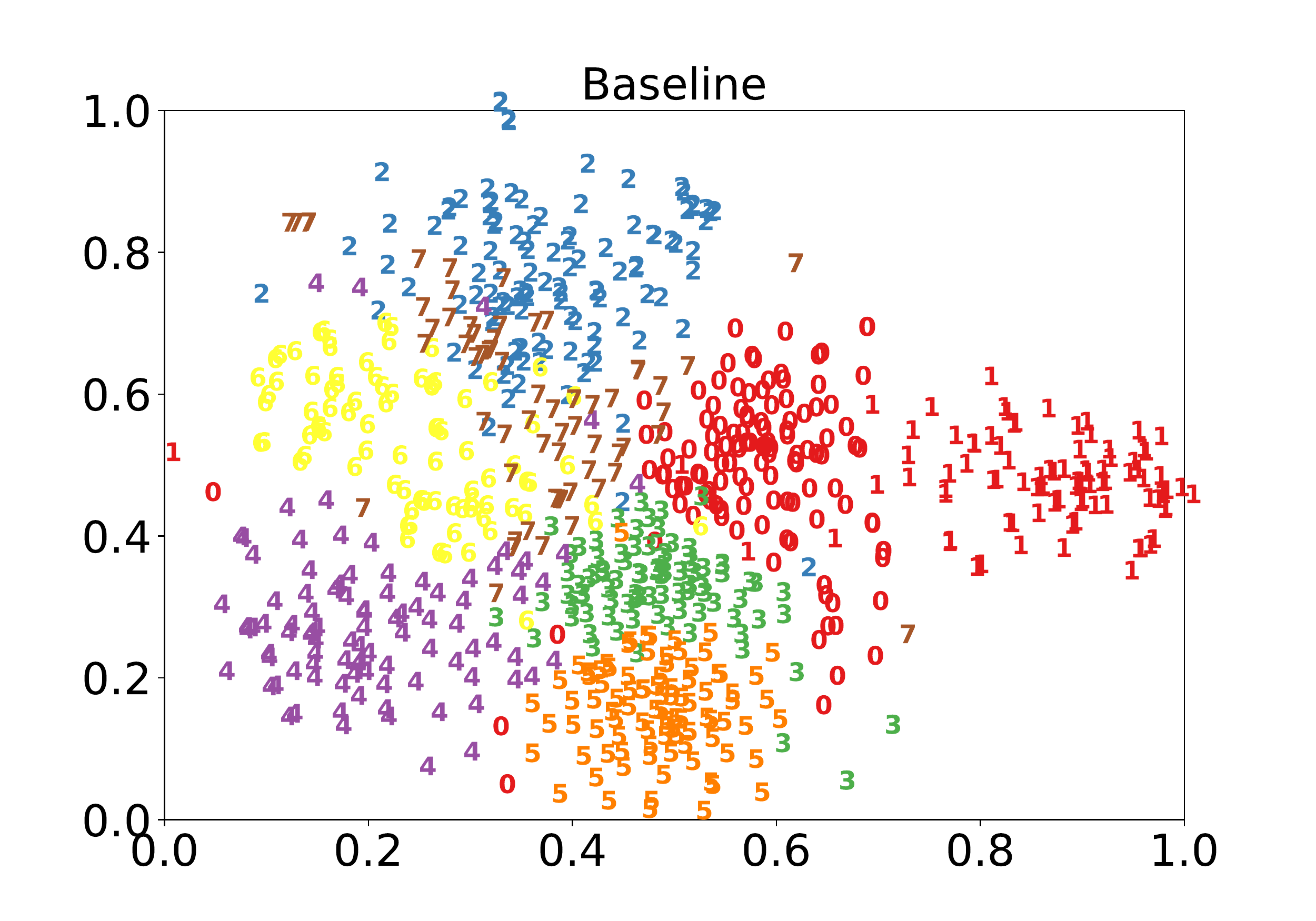}
\end{minipage}
\begin{minipage}{0.31\textwidth} 
 \includegraphics[width=2.5in]{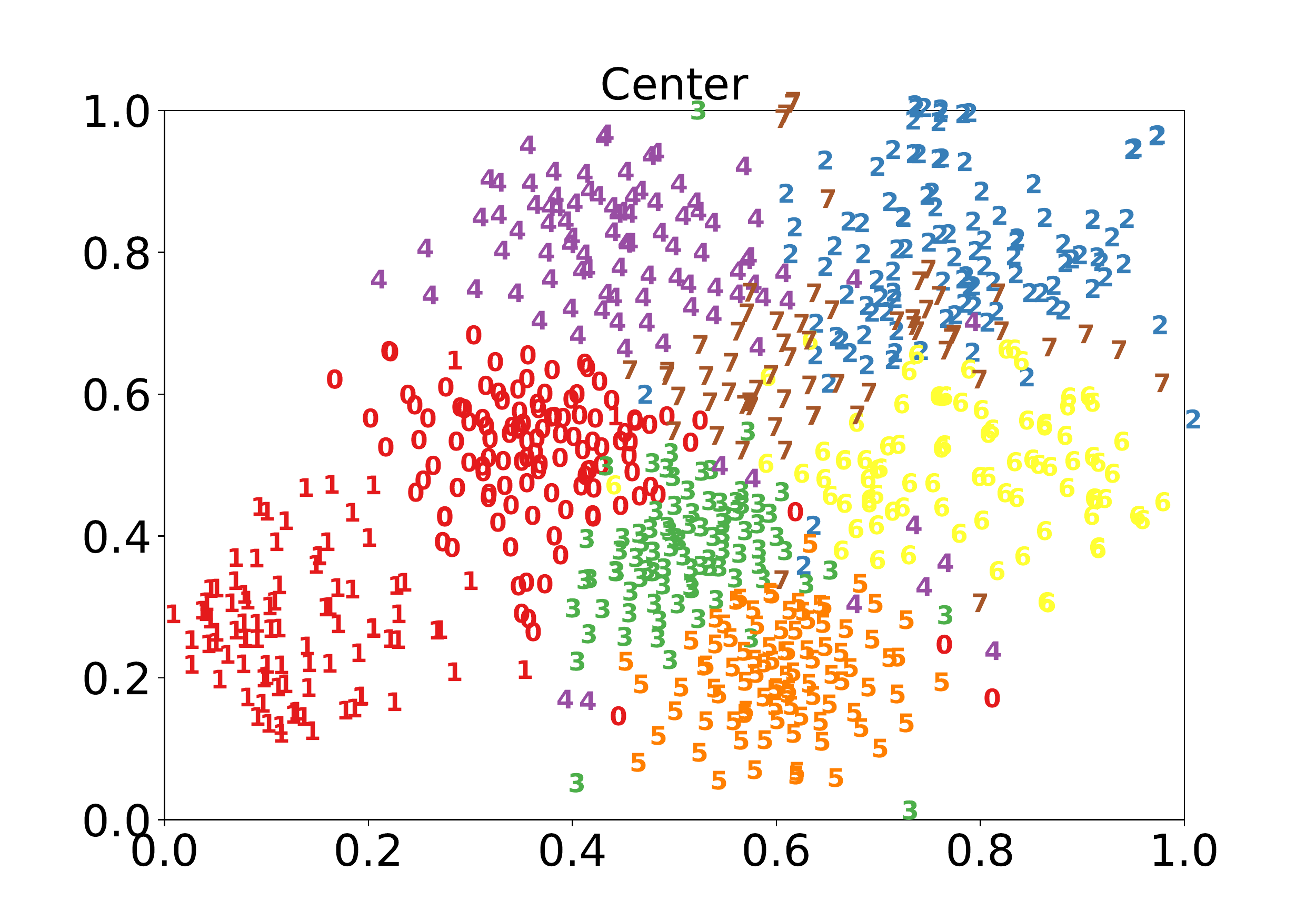}
\end{minipage}
\begin{minipage}{0.31\textwidth} 
 \includegraphics[width=2.5in]{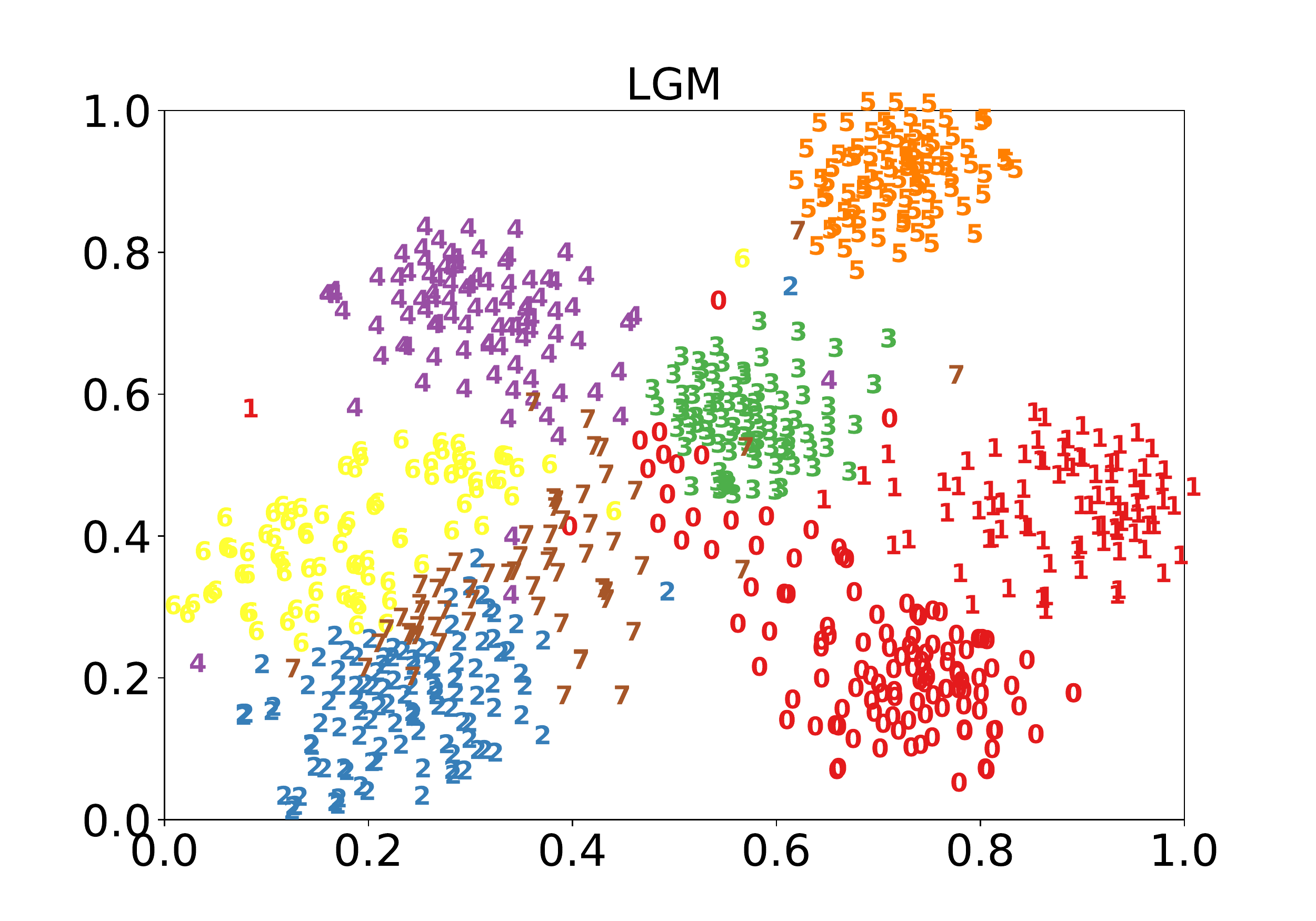}
\end{minipage}

\subfigure[t-SNE feature visualization on the training set] {
\begin{minipage}{0.31\textwidth} 
\includegraphics[width=2.5in]{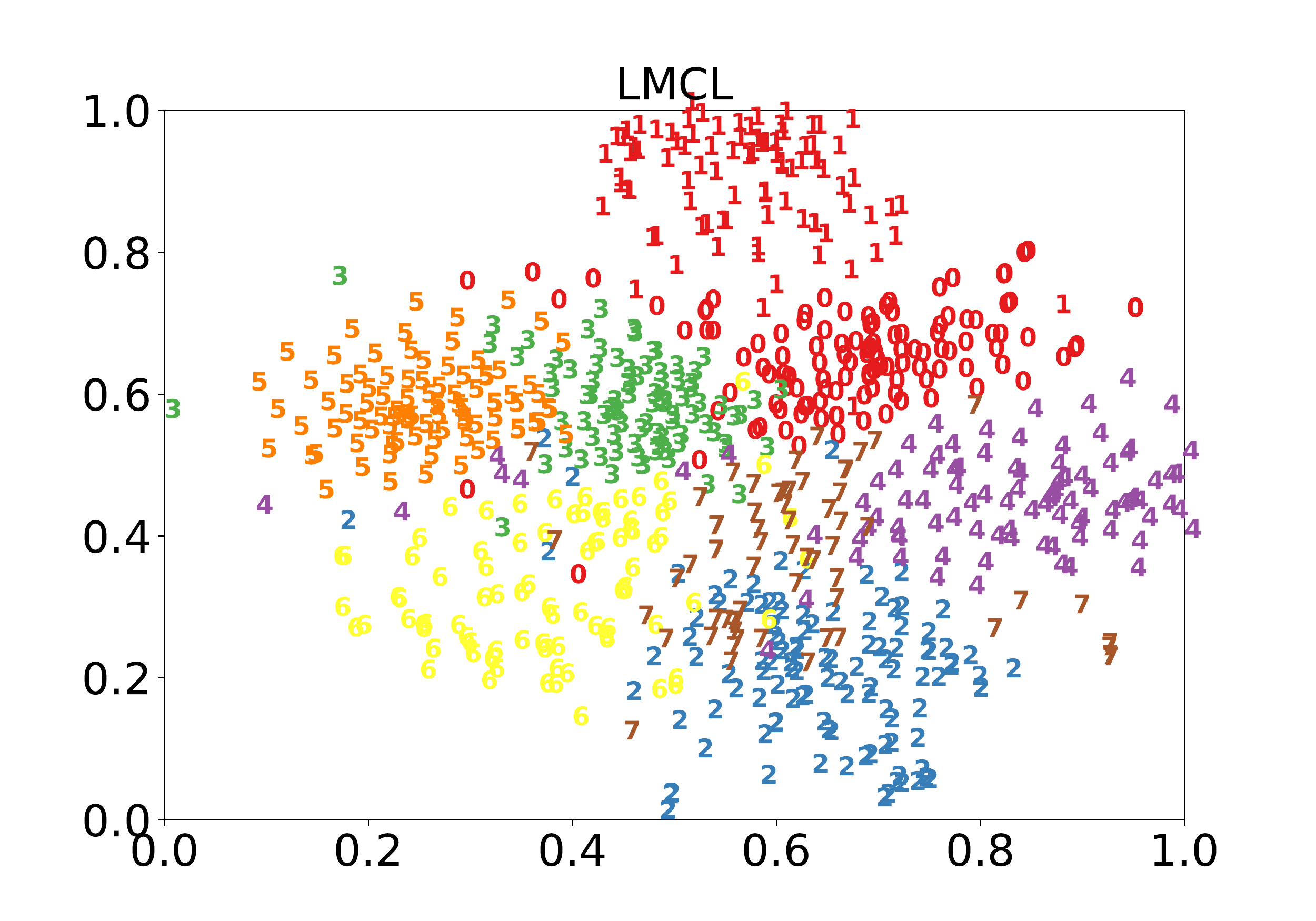}
\end{minipage}   
\begin{minipage}{0.31\textwidth} 
\includegraphics[width=2.5in]{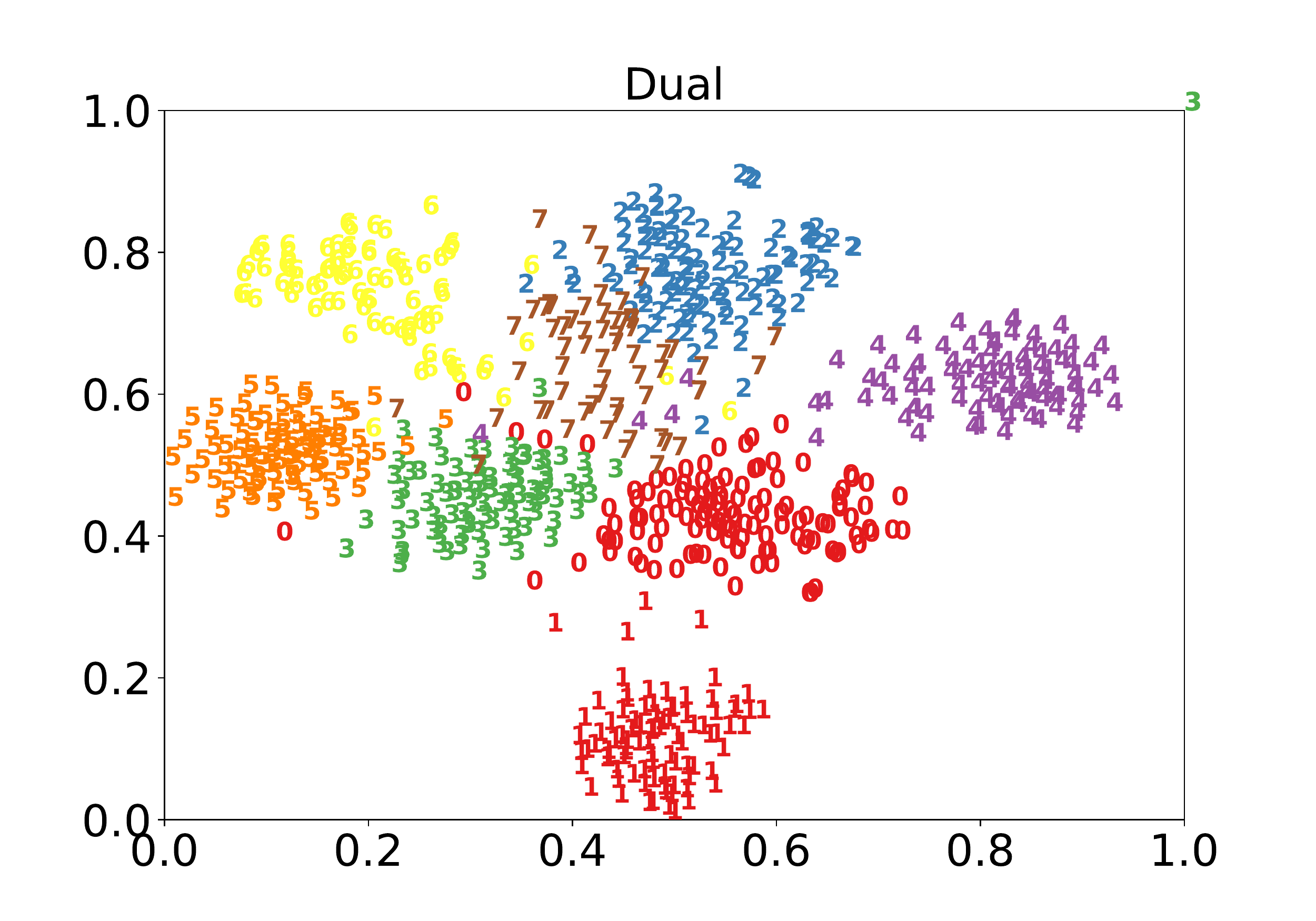}
\end{minipage}
\begin{minipage}{0.31\textwidth} 
\includegraphics[width=2.5in]{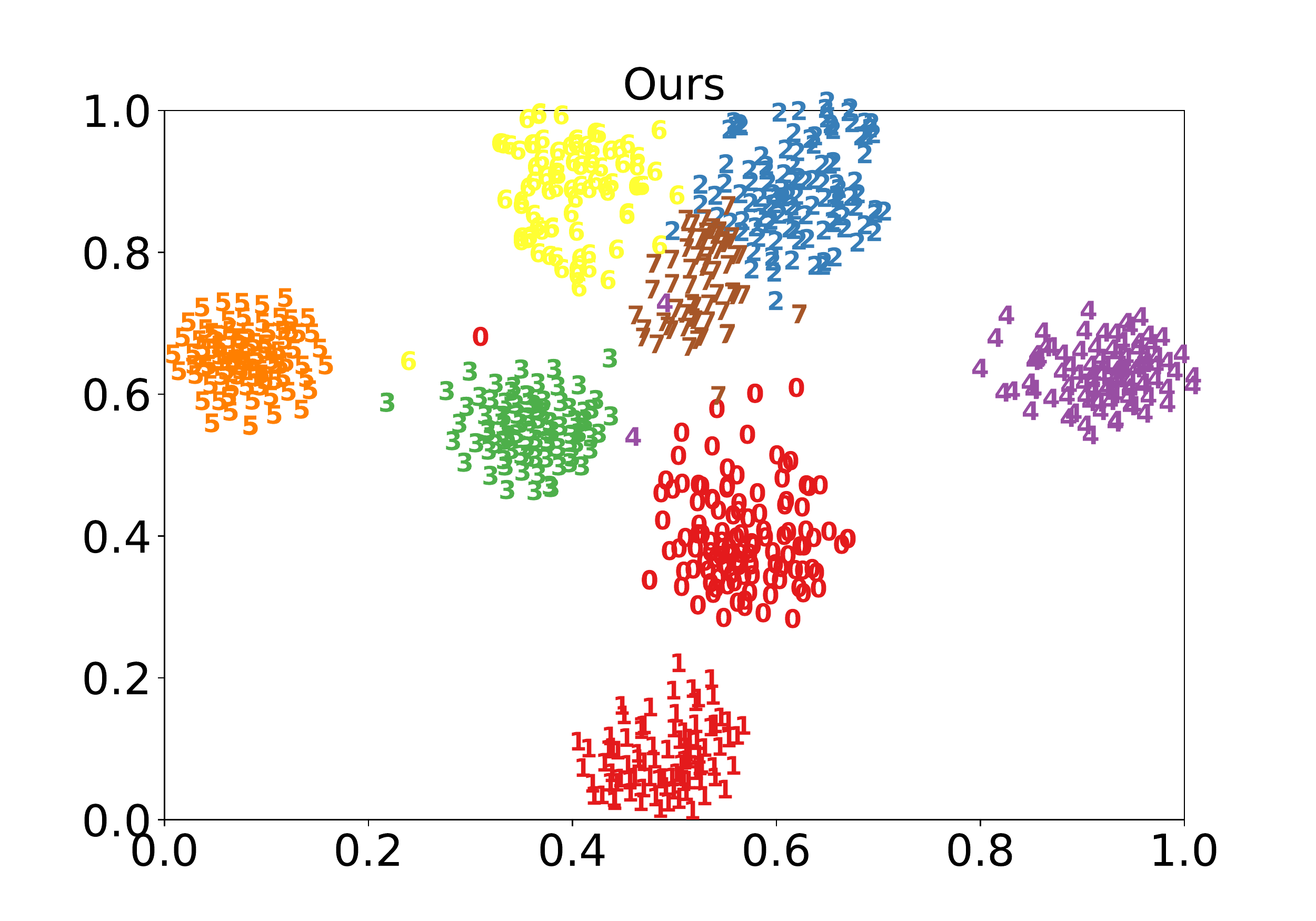}
\end{minipage}
}

\begin{minipage}{0.31\textwidth} 
\includegraphics[width=2.5in]{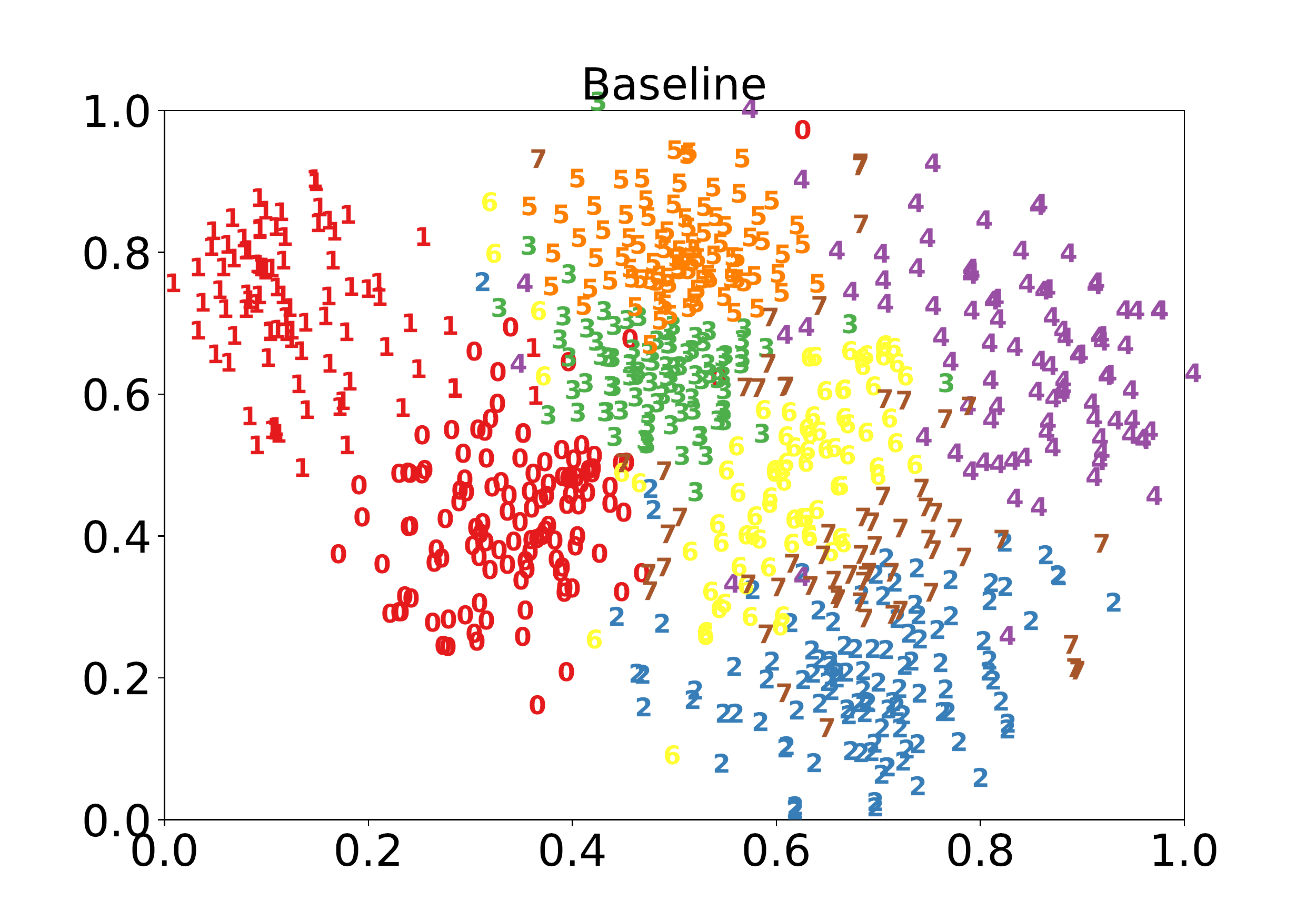}
\end{minipage}
\begin{minipage}{0.31\textwidth} 
\includegraphics[width=2.5in]{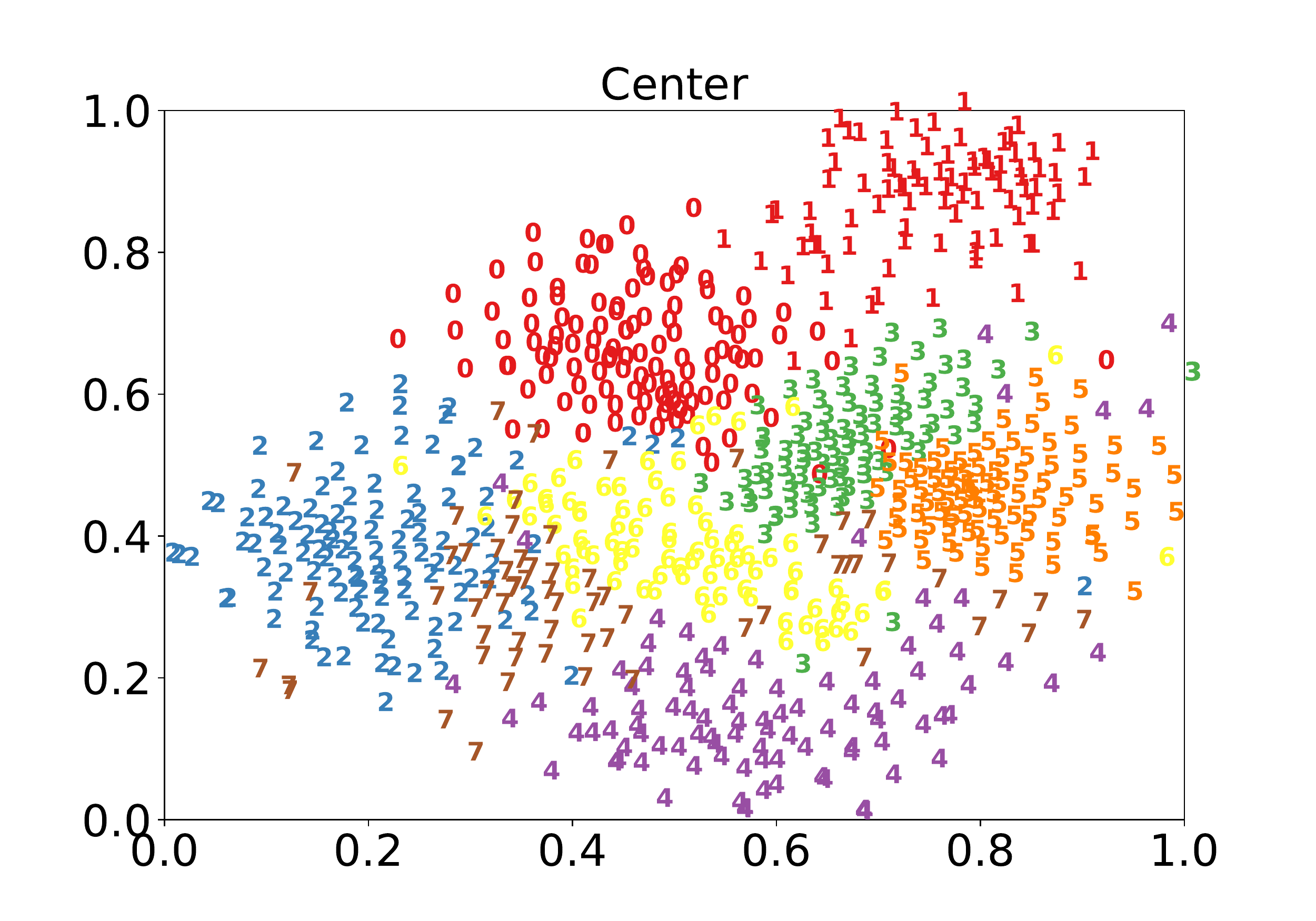}
\end{minipage}
\begin{minipage}{0.31\textwidth} 
\includegraphics[width=2.5in]{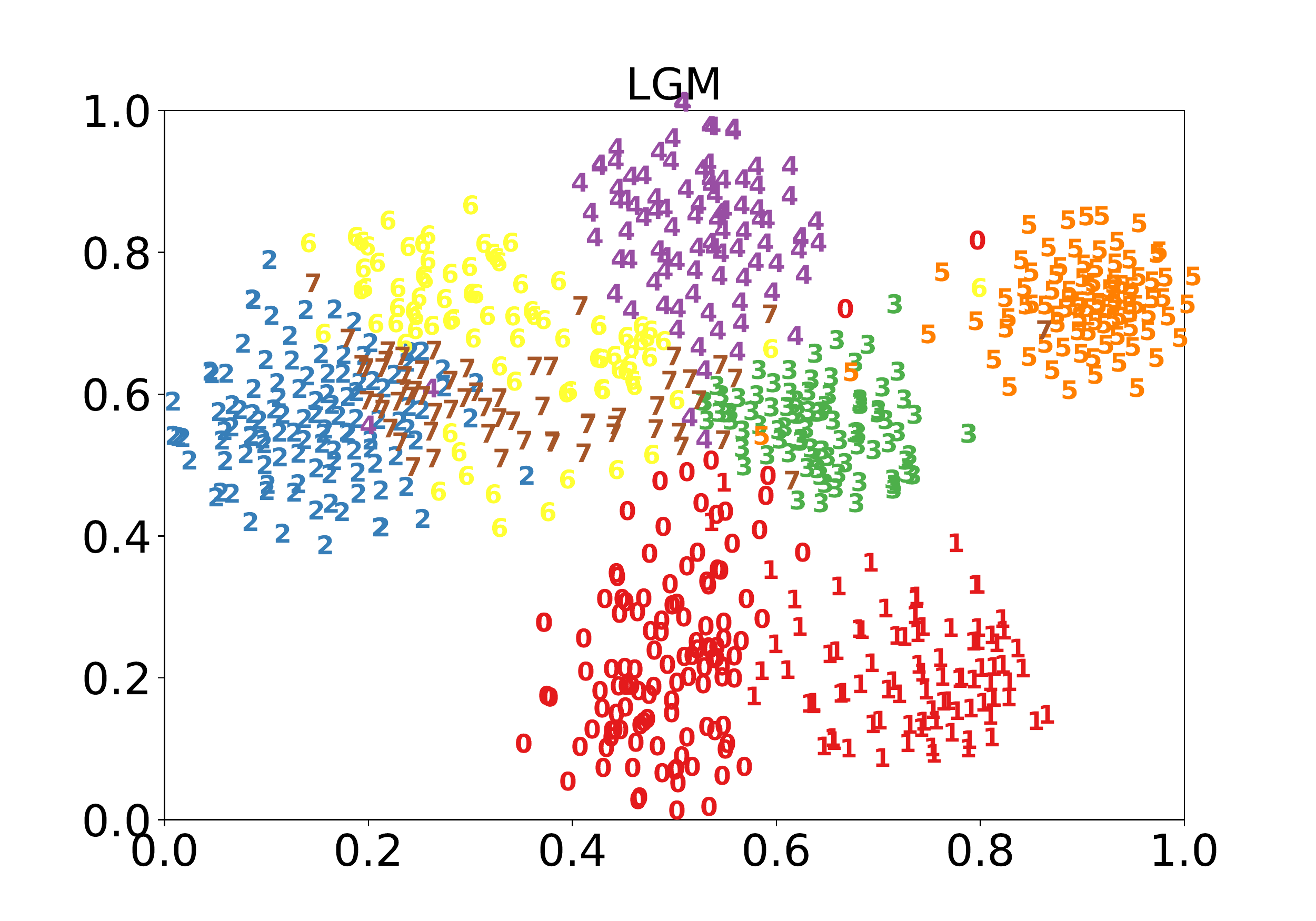}
\end{minipage}

\subfigure[t-SNE feature visualization on the test set] {
\begin{minipage}{0.31\textwidth} 
\includegraphics[width=2.5in]{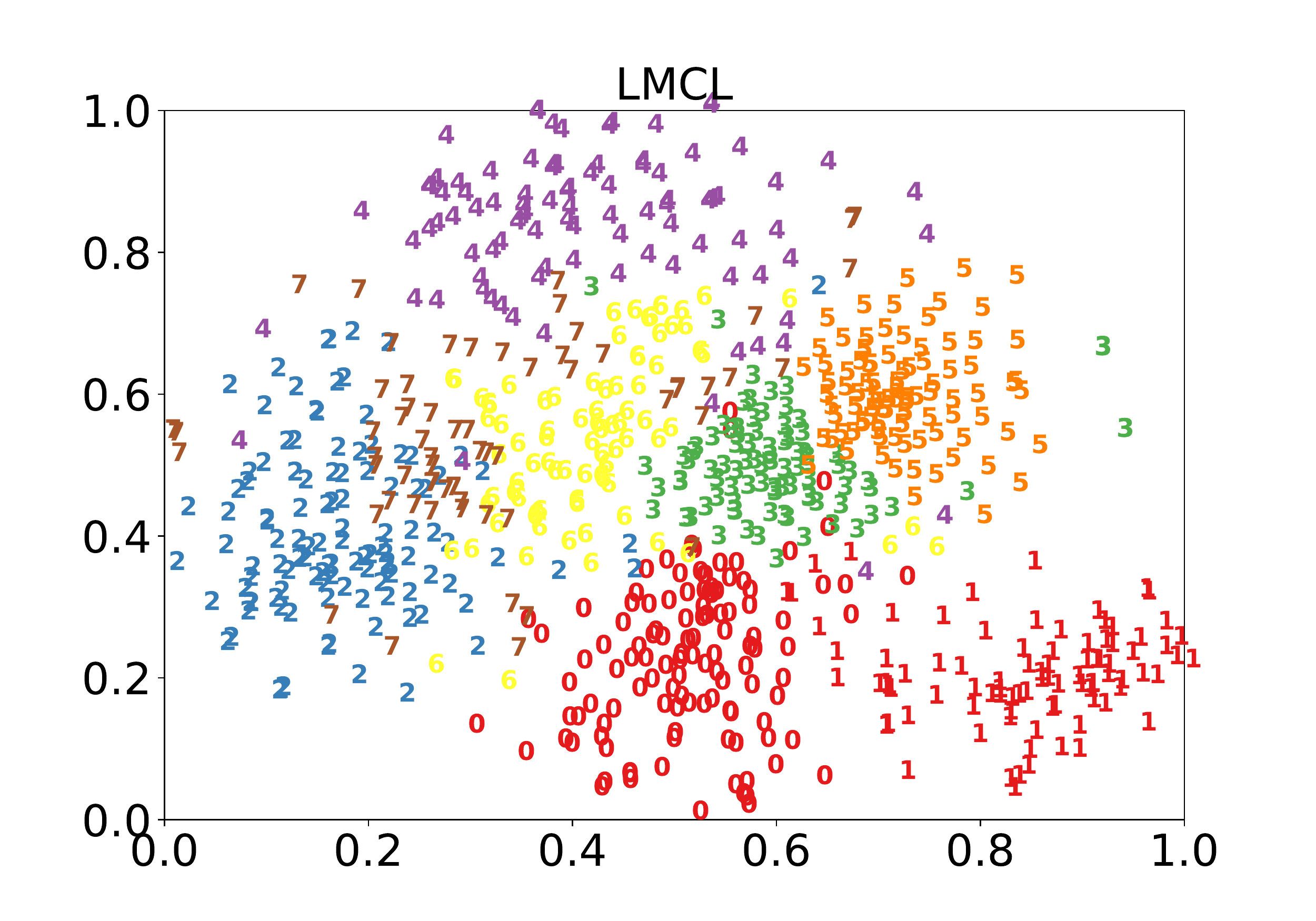}
\end{minipage}
\begin{minipage}{0.31\textwidth} 
\includegraphics[width=2.5in]{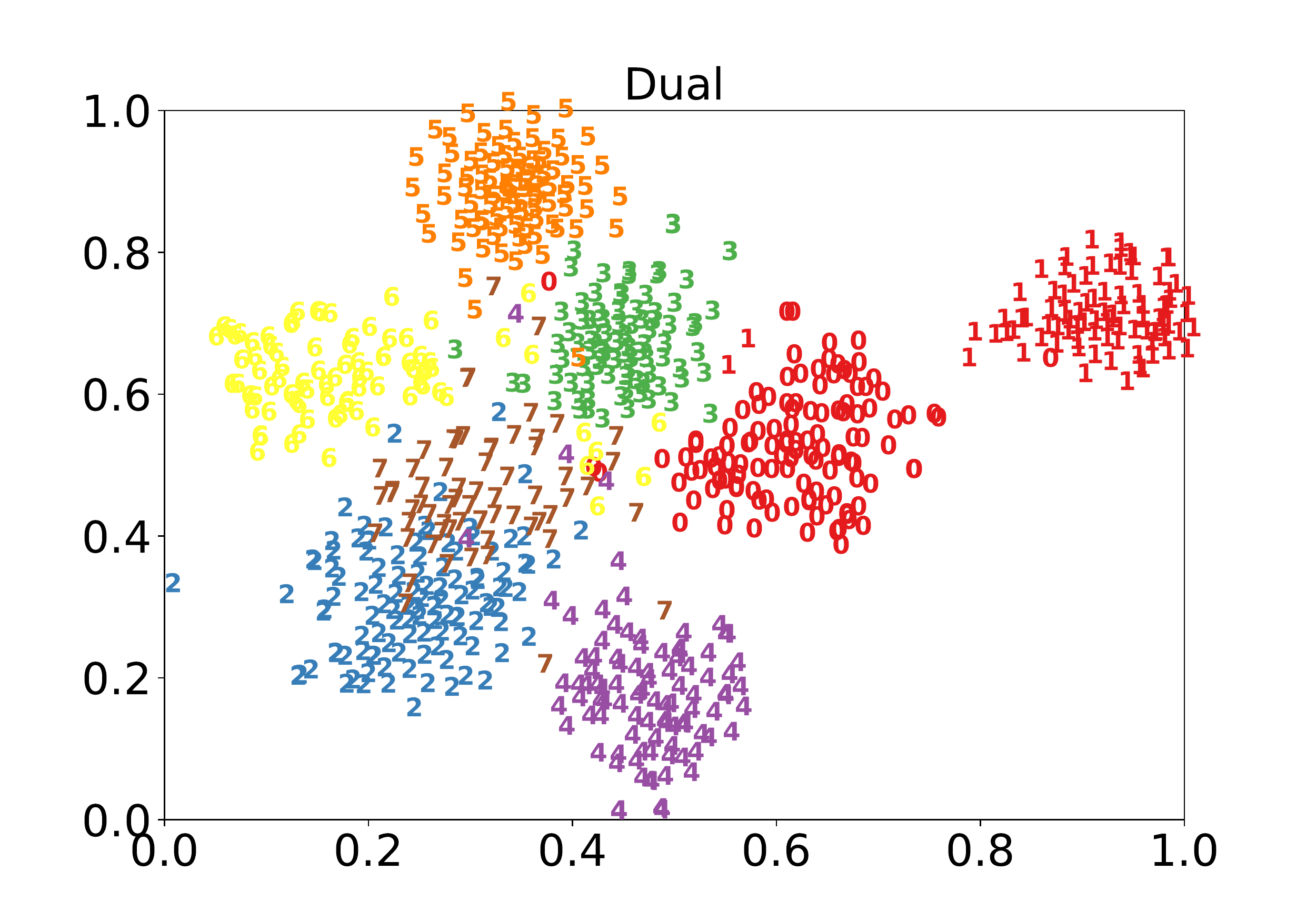}
\end{minipage}
\begin{minipage}{0.31\textwidth} 
\includegraphics[width=2.5in]{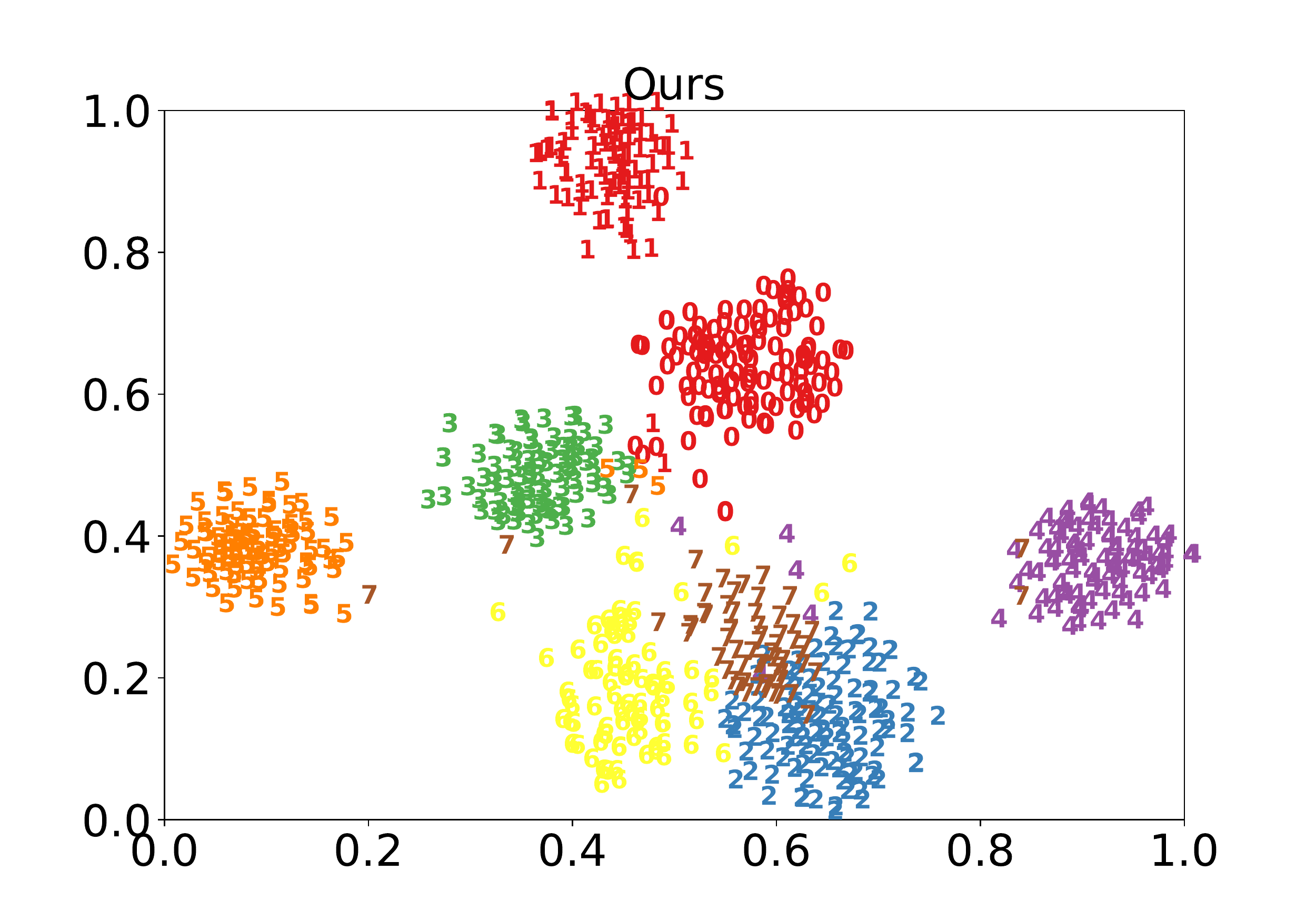}
\end{minipage}
}
\caption{Visualization of feature embeddings on the UIUC dataset.}\label{fig:tSNE}
\end{figure*}

We take the UIUC-Sport dataset as an example and visualize the features corresponding to different classes. We compare the features after training Baseline, Center, LGM, LMCL, Dual and the proposed ReMarNet. t-SNE~\cite{maaten2008visualizing} is used to depict the feature embedding $f_\varphi$ in two dimensions, and results are given in Figure~\ref{fig:tSNE}.

Figure~\ref{fig:tSNE}~(a) clearly shows that, on the training dataset, feature embeddings learned from ReMarNet are more compact for samples of the same class and more separated between samples of different classes, compared with other methods. Such a clear pattern can be seen on the test set as well, supporting the superior performance presented in Table~\ref{tab:sota}. While not presented, we observe similar patterns on other datasets as well. This confirms that our method is capable of learning discriminative features.

\subsection{Discussions}

Our experiments demonstrate that learning discriminative features on a small training set is indeed a challenging task for existing state-of-the-art methods. They cannot guarantee to improve classification accuracy over the baseline network trained with the cross entropy loss on all the four datasets, and their performance deteriorates more when the training set size gets further reduced. More specifically, Center and LMCL involve additional hyperparameters during training and the selected values may not generalize well to the test data. On top of this issue, LGM assumes a Gaussian mixture distribution on features, which may not always fit for the data. To generate a good performance from Snapshot, it is essential that the local minimum can be obtained within the given number of epochs and the model can escape the local minimum when restarting the optimization. In the setting of small sample size, these two requirements slightly contradict each other since a small learning rate is needed to fine-tune features produced from the feature extractor module and a large learning rate is needed to escape from the local minimum. 

Benefiting from a stricter requirement that the features should support a combined decision-making mechanism of two different classification paradigms, the proposed ReMarNet is capable of learning discriminative features and greatly enhances the baseline method on all of the evaluation datasets in this study. Its performance is also very competitive against state-of-the-art methods. 

While our method achieves better performance, it has slightly more learnable parameters than the baseline and other compared methods. When the backbone network is VGG16, all compared methods have 15.5M parameters and our network has 16.1M parameters. When the backbone network is set as AlexNet or Densenet-121, the numbers of parameters in all compared methods are 2.8M and 3.1M, respectively; the numbers of parameters in the ReMarNet are 8.6M and 9.8M, respectively.

\section{Conclusions, Limitations and Future Work}
In this paper, we propose a new deep neural network for small-sample image classification called \emph{Relation-and-Margin learning neural Network} (ReMarNet). It learns the discriminative features that can support both the classification paradigms based on the decision boundary and the similarity to class prototypes. Experimental results on four small datasets over a wide range of training sizes verify the efficacy of the proposed ReMarNet. 

Here we would like to share two ideas of future work. Firstly, the class prototypes are currently selected from the samples, and it would be more effective to learn more representative ones. Secondly, although the ReMarNet is proposed for image classification, its framework is quite generic; therefore, applying it to other data types such as text data would be another valuable future work.


\end{document}